\documentclass[10pt,twocolumn,letterpaper]{article}

\usepackage{cvpr}              % To produce the CAMERA-READY version
\usepackage{amstext} % for \text macro
\usepackage{array}   % for \newcolumntype macro
\newcolumntype{L}{>{$}l<{$}} % math-mode version of "l" column type
\usepackage{cuted}
\usepackage{amsmath}
\usepackage{amsfonts}
\usepackage{amssymb}
\usepackage{caption}
\usepackage{graphicx}
\usepackage{rotating}

\definecolor{cvprblue}{rgb}{0.21,0.49,0.74}
\usepackage[pagebackref,breaklinks,colorlinks,allcolors=cvprblue]{hyperref}

\def\paperID{81} % *** Enter the Paper ID here
\def\confName{CVPR}
\def\confYear{2026}
     
\title{NTIRE 2026 Rip Current Detection and Segmentation (RipDetSeg) Challenge Report}

\author{
Andrei Dumitriu$^{1, 2}$ \and  Aakash Ralhan$^{1}$ \and Florin Miron$^{2}$ \and Florin Tatui$^{2}$ \and  Radu Tudor Ionescu$^2$ \and Radu Timofte$^1$ \and Abdullah Naeem \and Anav Katwal \and Ayon Dey \and Md Tamjidul Hoque \and Asuka Shin \and Hiroto Shirono \and Kosuke Shigematsu \and Gaurav Mahesh \and Anjana Nanditha \and Jiji CV \and Akbarali Vakhitov \and Sang-Chul Lee \and Xinger Li \and Chun’an Yu \and Junhao Chen \and Yang Yang \and Gundluri Yuvateja Reddy \and Harshitha Palaram \and Gejalakshmi N \and Jeevitha S \and Jiachen Tu \and Guoyi Xu \and Yaoxin Jiang \and Jiajia Liu \and Yaokun Shi \and Amitabh Tripathi \and Modugumudi Mahesh \and Santosh Kumar Vipparthi \and Subrahmanyam Murala \and
$^{1}$Computer Vision Lab, CAIDAS \& IFI, University of Würzburg, Germany \\
$^{2}$University of Bucharest, Romania\\
{\tt\small {andrei.dumitriu}@uni-wuerzburg.de}\\
}

\begin{document}

\maketitle

\begin{abstract}
% This report presents the NTIRE 2026 Rip Current Detection and Segmentation (RipDetSeg) Challenge, which targets automatic rip current understanding in images. Rip currents are hazardous nearshore flows that remain difficult to identify because their visual appearance varies substantially across beaches, viewpoints, and sea states. To advance research on this safety-critical problem, the challenge builds on the RipVIS benchmark, evaluating both detection and segmentation. The dataset is diverse, containing images from more than $10$ countries, with $4$ camera orientations and diverse beach and sea conditions.

This report presents the NTIRE 2026 Rip Current Detection and Segmentation (RipDetSeg) Challenge, which targets automatic rip current understanding in images. Rip currents are hazardous nearshore flows that cause many beach-related fatalities worldwide, yet remain difficult to identify because their visual appearance varies substantially across beaches, viewpoints, and sea states. To advance research on this safety-critical problem, the challenge builds on the RipVIS benchmark, evaluating both detection and segmentation. The dataset is diverse, sourced from more than $10$ countries, with $4$ camera orientations and diverse beach and sea conditions. This report describes the dataset, challenge protocol, evaluation methodology, final results, and summarizes the main insights from the submitted methods. The challenge attracted $159$ registered participants and produced $9$ valid test submissions across the two tasks. Final rankings are based on a composite score that combines $F_1[50]$, $F_2[50]$, $F_1[40\!:\!95]$, and $F_2[40\!:\!95]$. Most participant solutions relied on pretrained models, combined with strong augmentation and post-processing design. These results suggest that rip current understanding benefits strongly from the robust general-purpose vision models' progress, while leaving ample room for future methods tailored to their unique visual structure.

\end{abstract}

{\let\thefootnote\relax\footnotetext{%
\hspace{-5mm} 
$^{1, 2}$
Andrei Dumitriu, Aakash Ralhan, 
Florin Miron, Florin Tatui,
Radu Tudor Ionescu and Radu Timofte are the CVPR 2026 NTIRE
Challenge organizers. The other authors participated in the challenge.
}}

\section{Introduction}
% from RipSeg

\begin{figure*}
\centering
\setlength{\tabcolsep}{1pt}
\begin{tabular}{c c c c}
     Aerial - Bird's Eye & Aerial - Tilted & Elevated Beachfront & Water-Level Beachfront \tabularnewline
     \begin{turn}{90} {\raggedright BBox} \end{turn}
     \includegraphics[width=0.22\textwidth]{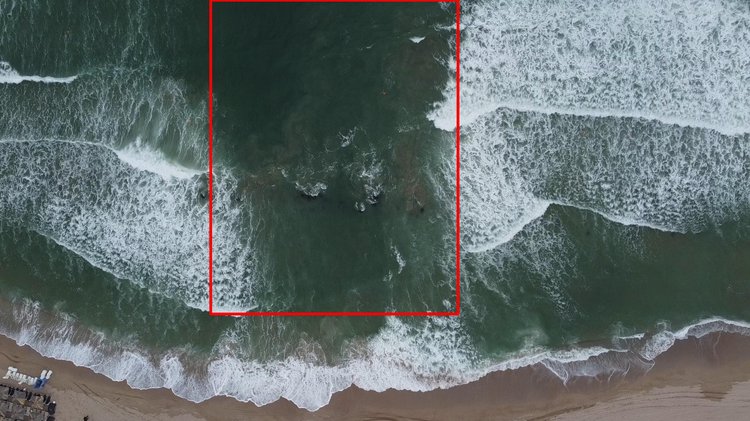} &
     \includegraphics[width=0.22\textwidth]{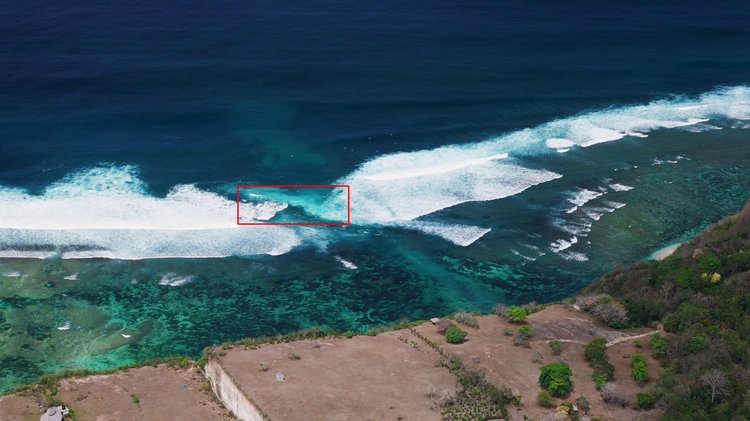} &
     \includegraphics[width=0.22\textwidth]{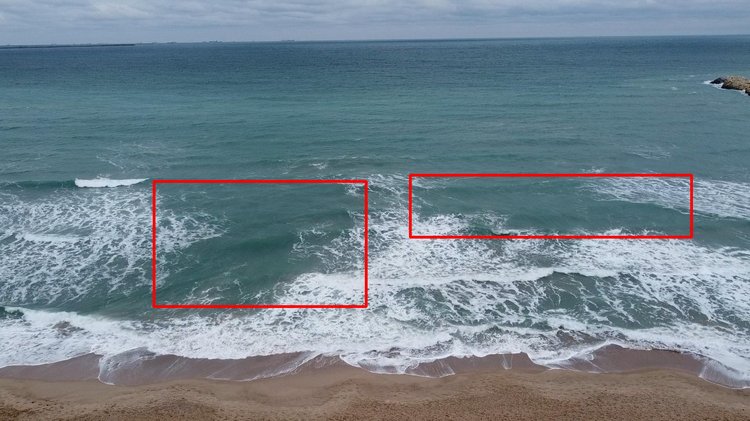} &
     \includegraphics[width=0.22\textwidth]{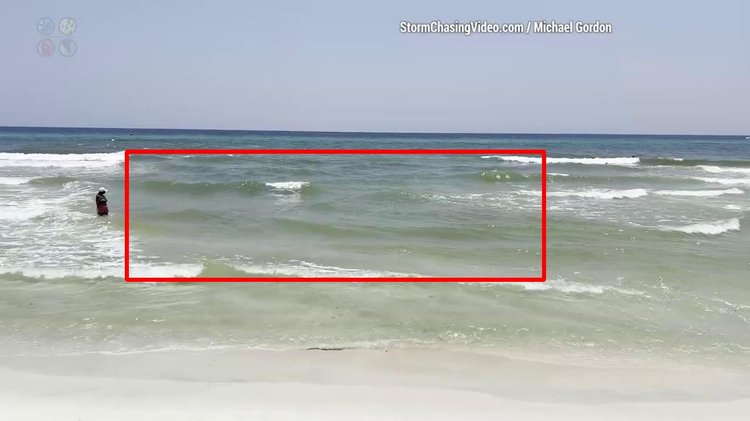}
     \tabularnewline
     \begin{turn}{90} {\raggedright BBox} \end{turn}
     \includegraphics[width=0.22\textwidth]{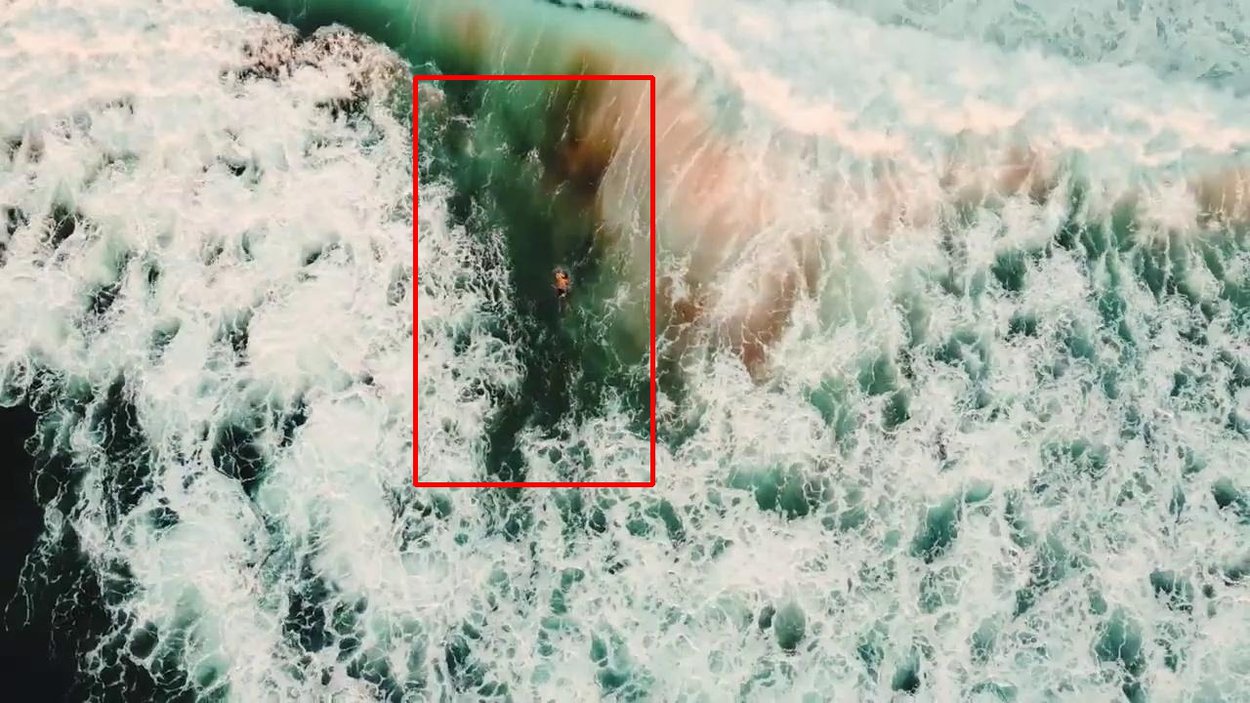} &
     \includegraphics[width=0.22\textwidth]{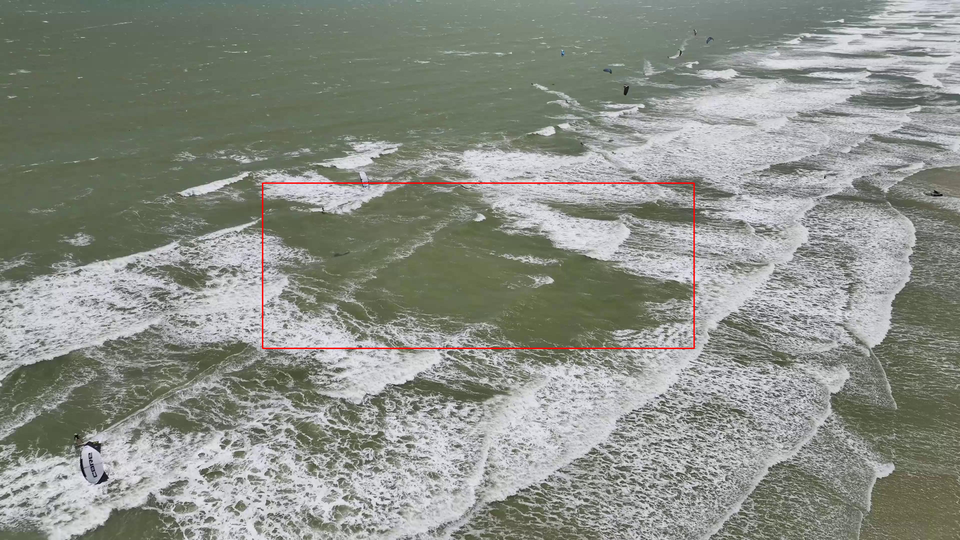} &
     \includegraphics[width=0.22\textwidth]{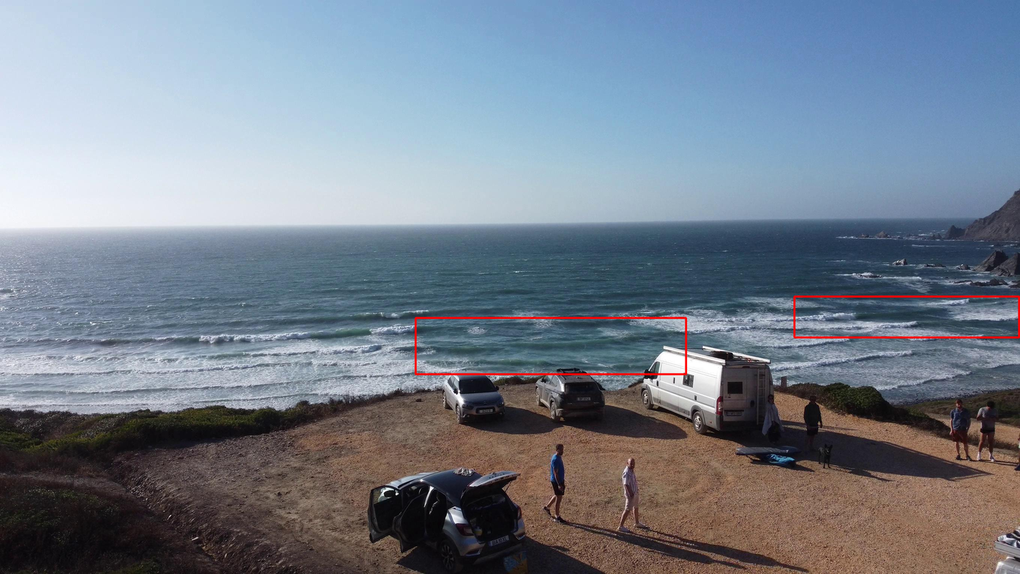} &
     \includegraphics[width=0.22\textwidth]{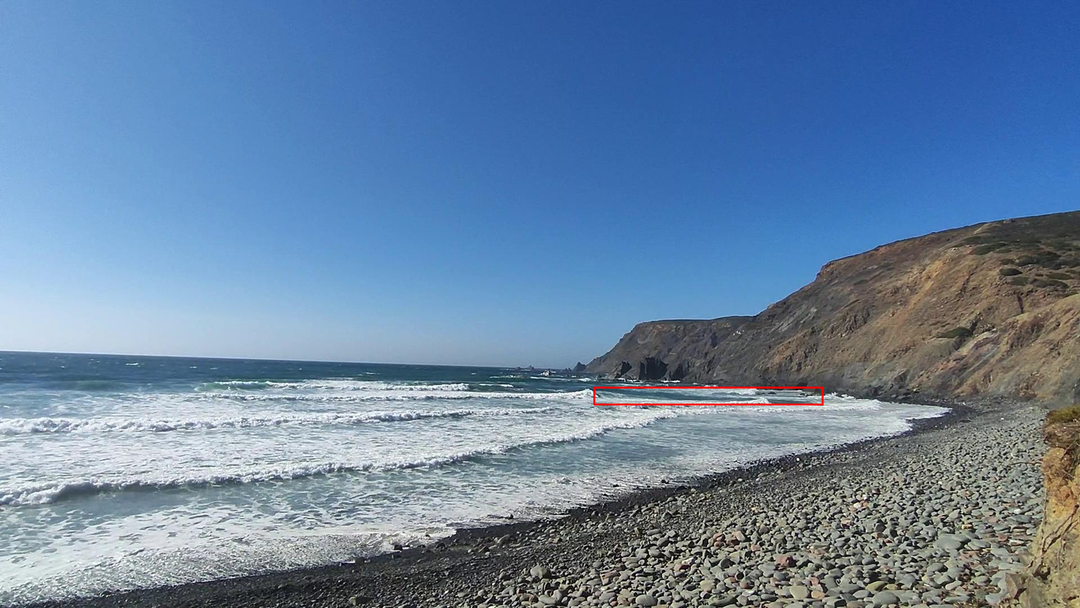}
     \tabularnewline
      \begin{turn}{90} {\raggedright Inst. Seg.} \end{turn}
     \includegraphics[width=0.22\textwidth]{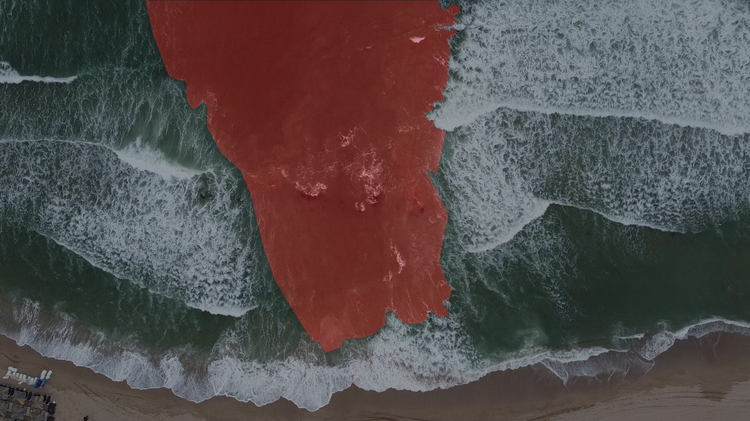} &
     \includegraphics[width=0.22\textwidth]{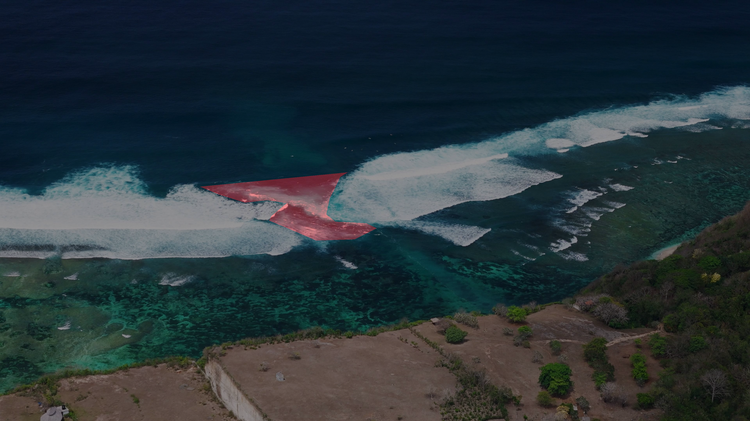} &
     \includegraphics[width=0.22\textwidth]{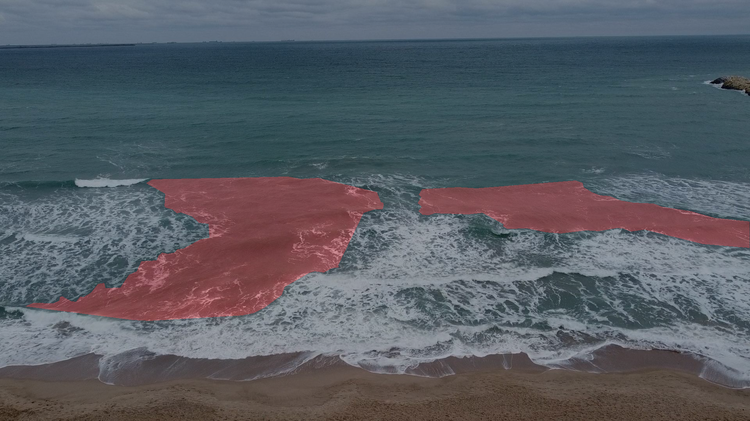} &
     \includegraphics[width=0.22\textwidth]{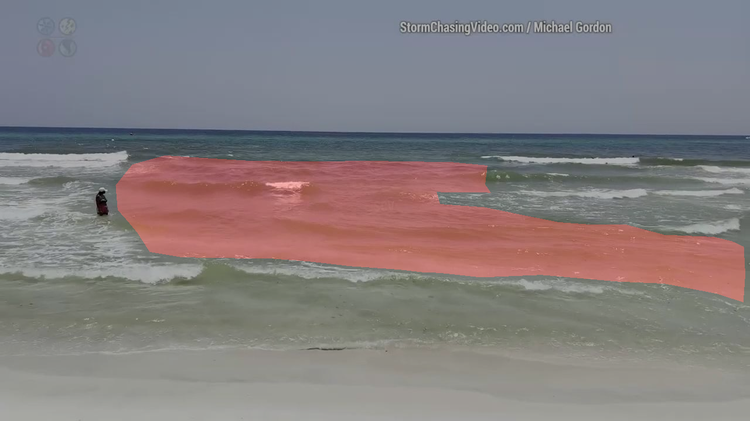}
     \tabularnewline
      \begin{turn}{90} {\raggedright Inst. Seg.} \end{turn}
     \includegraphics[width=0.22\textwidth]{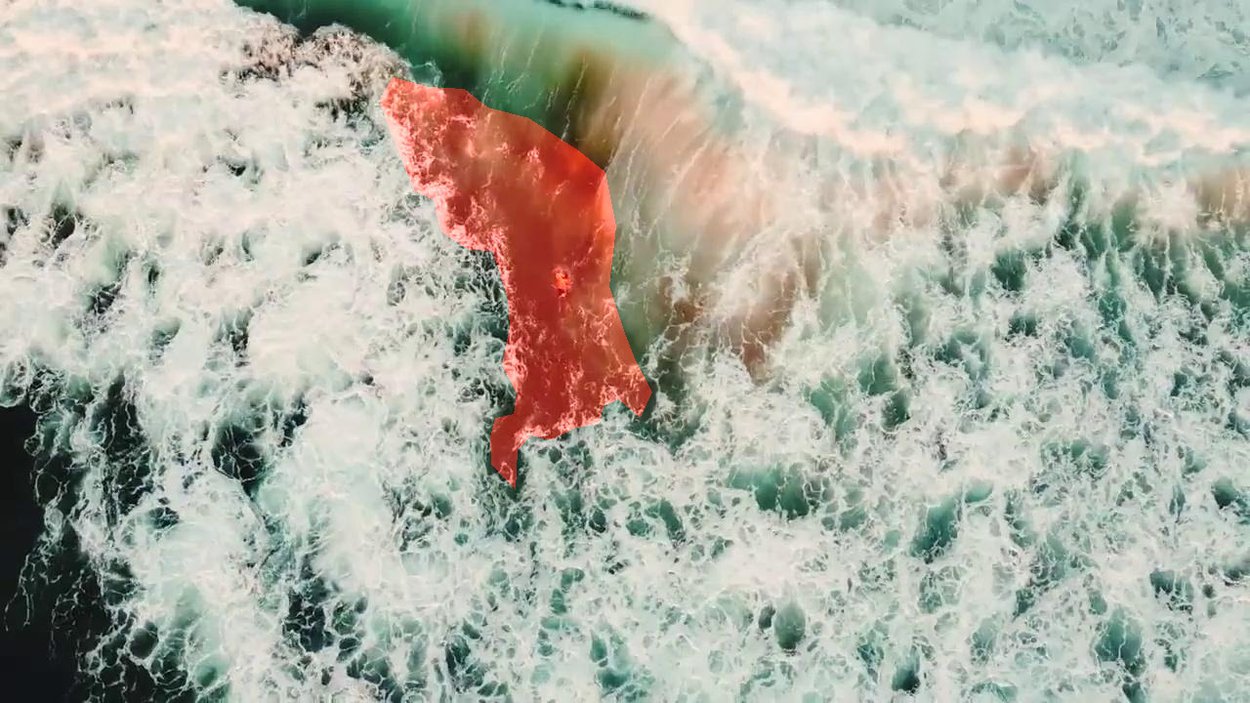} &
     \includegraphics[width=0.22\textwidth]{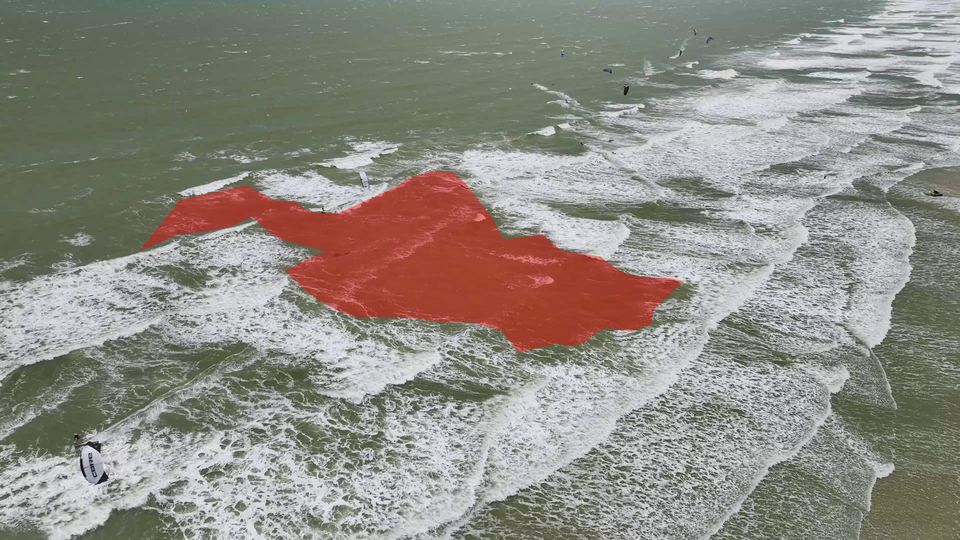} &
     \includegraphics[width=0.22\textwidth]{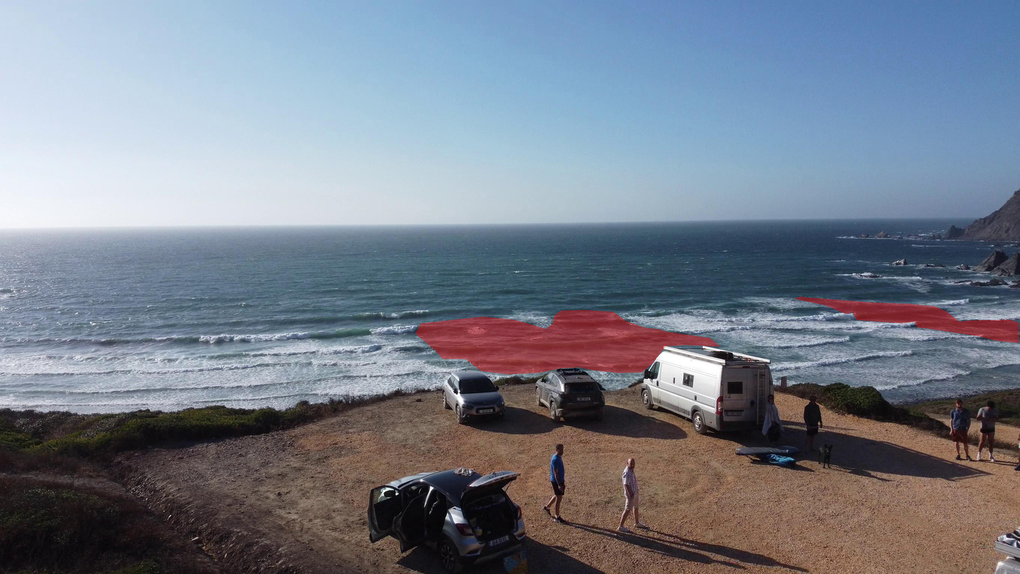} &
     \includegraphics[width=0.22\textwidth]{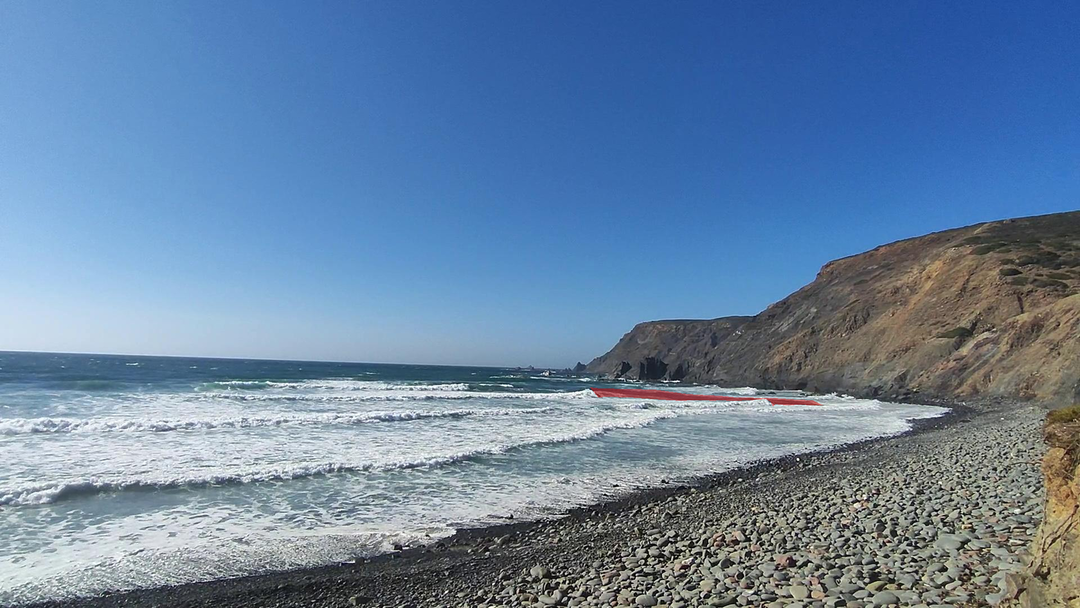}
     \tabularnewline
     \vspace{-0.4cm}
\end{tabular}
    \captionof{figure}{Examples from the RipVIS dataset \cite{Dumitriu_2025_CVPR}, which also forms the basis of the RipDetSeg Challenge. The four columns illustrate different camera orientations: (a) aerial bird’s-eye, (b) aerial tilted, (c) elevated beachfront, and (d) water-level beachfront. The examples highlight the diversity of rip currents across locations, types, and viewpoints. Rip currents are visible through disrupted wave-breaking patterns, sediment transport, and deflection flows, with annotations shown in red. Best viewed in color.}
\label{fig:intro_presentation}
\end{figure*}

Rip currents are strong, narrow seaward flows that transport water away from the shoreline, posing a significant hazard to swimmers and beachgoers worldwide \cite{da2003analysis, lushine1991study, brewster2019estimations, brander2013brief}. They occur along ocean, sea, and large lake coastlines, with their behavior governed by local hydrodynamics, seabed morphology, and, in some cases, anthropogenic coastal structures \cite{brander2000morphodynamics, castelle2016rip}. Rip currents can reach speeds exceeding $8.7$ km/h, faster than an Olympic swimmer, enabling them to quickly carry individuals offshore \cite{noaa2023ripcurrents}.

Despite their danger, rip currents are often difficult to identify visually, particularly for untrained observers. As a result, many swimmers instinctively attempt to swim against the current, leading to rapid exhaustion and increasing the risk of drowning. Although safety guidelines recommend escaping by swimming parallel to the shore, this strategy is only effective if the current is recognized in time. These challenges highlight the need for accurate and robust automatic detection systems, motivating the task addressed in this challenge.

Recent advances in computer vision have substantially improved performance in object detection, segmentation, and classification tasks \cite{he2017mask, yolov8, kirillov2023segment, ravi2024sam}, largely driven by the availability of large-scale, high-quality datasets such as COCO \cite{lin2014microsoft}, Cityscapes \cite{cordts2016cityscapes}, and YouTube-VIS \cite{Yang2019vis, vis2021}. In contrast, progress in automatic rip current identification has historically lagged behind, with earlier efforts being relatively fragmented and limited in scope. More recently, however, the topic has gained increasing attention within the computer vision community \cite{desilva2021automated, dumitriu2023rip, de2023ripviz, rashid2023reducing, zhu2022yolo, mori2022flow, mcgill2022flow, rampal2022interpretable, choi2024explainable, rashid2021ripdet, rashid2020ripnet, maryan2019machine, philip2016flow, 72_putri2025optimizing, 76_catala2026oriented, khan2025ripscout, khan2025ripfinder, 75_liu2026deep, 11_ali2025ibn, qian2025ripgan, 80_wan2026ripalert, aim2025ripseg, Dumitriu_2025_CVPR, 29_sun2023rip, 12_choi2025box2rip}.

\begin{table*}[t!]
    \centering
    \begin{tabular}{|c|ll|cccc|c|}
        \hline
        Rank & Team Name & Username &
        $F_1\uparrow$ & 
        $F_1[40\!:\!95]\uparrow$ & 
        $F_2\uparrow$ & 
        $F_2[40\!:\!95]\uparrow$ & 
        Final Score$\uparrow$ \\
        \hline
        1 & UNO Pixel Pros & anaeem & 69.26 & 41.25 & 70.61 & 42.06 & 55.79 \\
        2 & Riposte & gaurav\_mahesh & 67.55 & 41.92 & 61.26 & 38.02 & 52.19
        \\
        3 & SoloSeg & imdeali & 67.50 & 37.54 & 65.78 & 36.58 & 51.85
        \\
        4 & KMG & exllxe & 66.24 & 36.62 & 59.46 & 32.87 & 48.80 
        \\
        5 & RIP\_YuvatejaReddy & yuvatejareddy & 59.80 & 35.53 & 60.35 & 35.86 & 47.88 \\
        6 & SiGMoid & s2119 & 53.04 & 33.73 & 57.88 & 36.81 & 45.36 \\
        7 & NTR & miketjc0316 & 51.51 & 32.37 & 45.54 & 28.62 & 39.51 \\
        8 & Amitabh & amitabh & 44.37 & 27.66 & 47.41 & 29.56 & 37.25 \\
        \hline
    \end{tabular}
    \caption{\textbf{Segmentation task} quantitative results. The value for the final score is provided, alongside the values for $F_1$, $F_1[40\!:\!95]$, $F_2$, $F_2[40\!:\!95]$. The final score is computed as the average sum of the $4$ metrics. Results are computed on the NTIRE 2026 Rip Current Detection and Segmentation (RipDetSeg) test split.} % Add a caption if needed
    \label{tab:test_results_segmentation} % Add a label if needed
\end{table*}

\begin{table*}[t!]
    \centering
    \begin{tabular}{|c|ll|cccc|c|}
        \hline
        Rank & Team Name & Username &
        $F_1\uparrow$ & 
        $F_1[40\!:\!95]\uparrow$ & 
        $F_2\uparrow$ & 
        $F_2[40\!:\!95]\uparrow$ & 
        Final Score$\uparrow$ \\
        \hline
        1 & SiGMoid & s2119 & 67.86 & 46.65 & 65.67 & 45.15 & 56.33 \\
        2 & SoloSeg & imdeali & 68.45 & 45.12 & 66.72 & 43.97 & 56.06 \\
        3 & KMG & exllxe & 71.21 & 44.64 & 66.12 & 41.44 & 55.85 \\
        4 & Riposte & gaurav\_mahesh & 65.43 & 45.72 & 59.34 & 41.66 & 52.99 \\
        5 & RIP\_YuvatejaReddy & yuvatejareddy & 63.56 & 41.53 & 64.14 & 41.90 & 52.78 \\
        6 & VisionX & harsh1tha & 59.60 & 41.84 & 61.76 & 43.38 & 51.60 \\
        7 & NTR & miketjc0316 & 60.00 & 38.93 & 58.67 & 38.07 & 48.92 \\
        8 & Amitabh & amitabh & 48.52 & 32.46 & 51.84 & 34.68 & 41.87 \\
        \hline
    \end{tabular}
    \caption{\textbf{Detection task} quantitative results. The value for the final score is provided, alongside the values for $F_1$, $F_1[40\!:\!95]$, $F_2$, $F_2[40\!:\!95]$. The final score is computed as the average of $4$ metrics. Results are computed on the NTIRE 2026 Rip Current Detection and Segmentation (RipDetSeg) test split.} % Add a caption if needed
    \label{tab:test_results_detection} % Add a label if needed
\end{table*}

Despite this progress, automatic rip current detection remains a challenging problem. Unlike conventional objects, rip currents do not exhibit well-defined boundaries and instead manifest through diffuse, dynamic visual patterns. Their appearance varies significantly across different environmental conditions, camera viewpoints, and beach morphologies. Typical indicators, such as disruptions in wave structure, sediment plumes, or localized variations in water color, are often subtle, temporally unstable, and easily confounded by lighting changes, weather conditions, and perspective effects. These factors make robust detection and segmentation of rip currents particularly difficult, motivating the need for specialized datasets and evaluation frameworks such as the one introduced in this challenge.

The introduction of RipVIS \cite{Dumitriu_2025_CVPR}, the largest publicly available dataset for rip current analysis, represented a significant step forward by establishing a large-scale benchmark for video instance segmentation. RipVIS initially focused on a single task: instance segmentation, which is particularly well-suited for capturing the complex, amorphous structure of rip currents. Unlike classification, which provides no spatial localization, or bounding box detection, which may include substantial amounts of irrelevant background, instance segmentation enables precise delineation of the full spatial extent of rip currents without omitting critical regions.

Detection can still serve as a practical alternative. It is more established in the community, easier to implement, and simpler to integrate into existing pipelines. In settings where detection models may be more reliable than segmentation models, detection can still be the more useful choice, as its coarse localization is often sufficient for early warning and monitoring applications, even though it provides less precise spatial detail. Together, these perspectives motivate the inclusion of both detection and instance segmentation, enabling a more comprehensive evaluation of methods across different accuracy-efficiency trade-offs relevant to practical beach safety systems.

Building on RipVIS and the RipSeg Challenge at ICCV 2025 AIM workshop \cite{aim2025ripseg}, we introduce the RipDetSeg Challenge at CVPR 2026 NTIRE workshop, extending the task to include both instance segmentation and object detection of rip currents in still images. While preserving the core problem formulation of rip current recognition under complex coastal conditions, RipDetSeg broadens the scope by requiring models to perform both pixel-level delineation and bounding box localization, reflecting complementary requirements in real-world beach safety systems. The resulting dataset contains a large and diverse set of scenes, covering a wide range of environmental conditions, camera viewpoints, and coastal morphologies, as seen in Figure \ref{fig:intro_presentation}. As in the previous edition, the data is split into training, validation, and held-out test sets, with strict separation to ensure fair evaluation and to discourage overfitting.

The challenge remains focused on single-class detection and segmentation, where multiple rip current instances may be present in a single image. By jointly evaluating detection and segmentation performance, RipDetSeg encourages the development of models that can both localize and precisely delineate rip currents, addressing the dual requirements of interpretability and actionable accuracy. As in RipSeg, we adopt a custom evaluation metric that combines standard performance measures with application-driven priorities, emphasizing robustness and recall in safety-critical scenarios.

%%% end from RipSeg

This challenge is one of the challenges associated with the NTIRE 2026 Workshop\footnote{\url{https://www.cvlai.net/ntire/2026/}} on: deepfake detection~\cite{ntire26deepfake}, 
high-resolution depth~\cite{ntire26hrdepth}, multi-exposure image fusion~\cite{ntire26raim_fusion}, 
AI flash portrait~\cite{ntire26raim_portrait}, professional image quality assessment~\cite{ntire26raim_piqa}, light field super-resolution~\cite{ntire26lightsr}, 3D content super-resolution~\cite{ntire263dsr}, bitstream-corrupted video restoration~\cite{ntire26videores}, X-AIGC quality assessment~\cite{ntire26XAIGCqa}, shadow removal~\cite{ntire26shadow}, ambient lighting normalization~\cite{ntire26lightnorm}, controllable Bokeh rendering~\cite{ntire26bokeh}, rip current detection and segmentation~\cite{ntire26ripdetseg}, low light image enhancement~\cite{ntire26llie}, high FPS video frame interpolation~\cite{ntire26highfps}, Night-time dehazing~\cite{ntire26nthaze,ntire26nthaze_rep}, learned ISP with unpaired data~\cite{ntire26isp}, short-form UGC video restoration~\cite{ntire26ugcvideo}, raindrop removal for dual-focused images~\cite{ntire26dual_focus}, image super-resolution (x4)~\cite{ntire26srx4}, photography retouching transfer~\cite{ntire26retouching}, mobile real-word super-resolution~\cite{ntire26rwsr}, remote sensing infrared super-resolution~\cite{ntire26rsirsr}, AI-Generated image detection~\cite{ntire26aigendet}, cross-domain few-shot object detection~\cite{ntire26cdfsod}, financial receipt restoration and reasoning~\cite{ntire26finrec}, real-world face restoration~\cite{ntire26faceres}, reflection removal~\cite{ntire26reflection}, anomaly detection of face enhancement~\cite{ntire26anomalydet},
video saliency prediction~\cite{ntire26videosal}, efficient super-resolution~\cite{ntire26effsr},
3d restoration and reconstruction in adverse conditions~\cite{ntire26realx3d}, image denoising~\cite{ntire26denoising}, blind computational aberration correction~\cite{ntire26aberration},
event-based image deblurring~\cite{ntire26eventblurr}, efficient burst HDR and restoration~\cite{ntire26bursthdr}, low-light enhancement: `twilight cowboy'~\cite{ntire26twilight}, and efficient low light image enhancement~\cite{ntire26effllie}.

\begin{table*}[t!]
\centering
\renewcommand{\arraystretch}{1.1}
\setlength{\tabcolsep}{3pt}
\begin{tabular}{|p{3.3cm}|p{4.7cm}|p{5.5cm}|p{3cm}|}
\hline
{Team} & {Members} & {Affiliations} & Contact\\
\hline

NTIRE 2026 Rip Current Segmentation (RipDetSeg) & 
Andrei Dumitriu$^{1,2}$, Aakash Ralhan$^{1}$, Florin Miron$^{2}$, Florin Tatui$^{2}$, Radu Tudor Ionescu$^{2}$, Radu Timofte$^{1}$&
$^1$ Computer Vision Lab, IFI \& CAIDAS, University of W\"urzburg, Germany \newline
$^2$ University of Bucharest, Romania & 
andrei.dumitriu@uni-wuerzburg.de \\
\hline
UNO Pixel Pros &
Abdullah Naeem$^{1}$, Anav Katwal$^{1}$, Ayon Dey$^{1}$, Md Tamjidul Hoque$^{1}$ &
$^1$ University of New Orleans, USA & 
anaeem@uno.edu \\
\hline
SiGMoid &
Asuka Shin$^{1}$, Hiroto Shirono$^{2}$, Kosuke Shigematsu$^{1}$ &
$^1$National Institute of Technology, Oita College, Japan \newline
$^2$Kyushu Institute of Technology, Japan &
jasasu602@ gmail.com \\
\hline
Riposte &
Gaurav Mahesh$^{1}$, Anjana Nanditha$^{1}$, Jiji CV$^{1}$ &
$^1$Shiv Nadar University, Chennai, India &
gaurav.maheshh@ gmail.com \\
\hline
SoloSeg &
Akbarali Vakhitov$^{1}$, Sang-Chul Lee$^{1}$ &
$^1$Inha University, South Korea &
vahidovakbarali @gmail.com \\
\hline
KMG &
Xinger Li${^1}$, Chun’an Yu${^1}$, Junhao Chen${^1}$, Yang Yang${^1}$ &
$^1$Nanjing University of Science and Technology, China &
865069924@qq.com \\
\hline
RIP\_YuvatejaReddy &
Gundluri Yuvateja Reddy${^1}$ &
$^1$Shiv Nadar University Chennai, India &
yuvatejareddygundl- uru@gmail.com \\
\hline
VisionX &
Harshitha Palaram${^1}$, Gejalakshmi N${^1}$, Jeevitha S${^1}$ &
$^1$Shiv Nadar University Chennai, India &
harshithapalaram09 @gmail.com \\
\hline 
NTR &
Jiachen Tu${^1}$, Guoyi Xu${^1}$, Yaoxin Jiang${^1}$, Jiajia Liu${^1}$, Yaokun Shi${^1}$ &
$^1$University of Illinois Urbana-Champaign, USA &
jtu9@illinois.edu \\
\hline
Amitabh &
Amitabh Tripathi${^1}$, Modugumudi Mahesh$^{1}$, Santosh Kumar Vipparthi${^1}$, Subrahmanyam Murala${^2}$ &
$^1$IIT Ropar, India \newline
$^2$Trinity College Dublin, Ireland &
amitabh.23eez0025 @iitrpr.ac.in \\
\hline

\end{tabular}
\caption{Teams, members, and affiliations for NTIRE 2026 Rip Current Detection and Segmentation (RipDetSeg) Challenge.}
\label{tab:teams_affil}
\end{table*}

\section{Challenge format and ranking}
The NTIRE 2026 Rip Current Detection and Segmentation (RipDetSeg) Challenge was organized on CodaBench \cite{codabench} and comprised two tasks, detection and segmentation, which were conducted over three phases, named intro, validation, and test. In the first phase (intro), we popularized the challenge and introduced the tasks and relevant papers, allowing all users some time to prepare in advance. In the second phase (validation), users were provided with a set of $32,407$ training images, alongside polygon and axis-aligned bounding box annotations, serving as training data. They were also provided with an additional $9,262$ images, alongside a COCO JSON without annotations, to ensure correct prediction format. The goal was to submit their predictions on the CodaBench server and see their results when compared to the ground truth. The submission system allowed for a maximum $10$ submissions per day, for either detection or segmentation task, or both at the same time, depending on interest. The participants had six weeks for this phase, during which they could develop and fine-tune their models. In the last phase (test), participants were provided $9,145$ images without annotations, and they had one day to submit their results and see their final score. The short window of time was intentionally designed to minimize the risk of fraud. In both validation and test phases, the file names were randomized in order not to facilitate inferring the structure of the names. As a final verification, only teams that passed the reproducibility test were included in the final ranking, presented in Table \ref{tab:test_results_segmentation} for segmentation and Table \ref{tab:test_results_detection} for detection.

Correctly identifying rip currents is a safety-critical task, where a false positive can be a nuisance while a false negative can potentially be deadly. Therefore, we encourage the use of the $F_2$ score when evaluating rip current detection for real-world scenarios, to minimize false negatives. For this challenge, we used a composite score of $F_1$ and $F_2$, evaluated at different IoU thresholds ($F_1[50]$, $F_1[40\!:\!95]$, $F_2[50]$, $F_2[40\!:\!95]$) to balance precision and recall. In this notation, $[50]$ refers to evaluation at an IoU threshold of $0.50$, whereas $[40\!:\!95]$ represents the average score over thresholds ranging from $0.40$ to $0.95$ with a step size of $0.05$. The final score is defined in Eq.~\eqref{eq:final_score}, combining the assessment of $F_1$ with the recall-oriented emphasis of $F_2$. In this way, the metric reasonably reflects real-world usefulness while also promoting proper model design for the challenge.

\begin{equation}
\label{eq:final_score}
\texttt{score}  = \frac{F_1[50]+F_1[40\!:\!95]+F_2[50] +F_2[40\!:\!95]}{4}
\end{equation}

% \usepackage{xcolor}

% \usetikzlibrary{
%     positioning,
%     arrows.meta,
%     fit,
%     backgrounds
% }

\section{Methods}
The affiliations of challenge organizers and participants are
included in Table \ref{tab:teams_affil}. In this section, we briefly present the approach submitted by each team.

\begin{figure}[t]
    \centering
    \includegraphics[width=0.95\columnwidth]{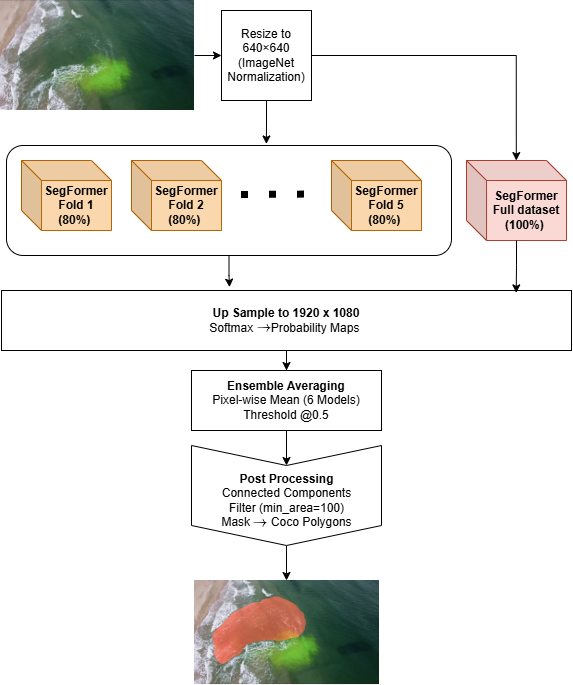}
    \caption{Segmentation pipeline showing Team UNO Pixel Pros's 6-model ensemble approach. Five models trained via cross-validation and one on the whole training dataset generate predictions that are averaged pixel-wise, thresholded ($0.5$), converted into instance-level masks using connected components analysis, and filtered (min area $100$).}
    \label{fig:uno_pixel_pros}
\end{figure}

\subsection{Team UNO Pixel Pros}
% Team UNO Pixel Pros used a SegFormer-B5~\cite{xie2021segformer}  segmentation model with a 6-model ensemble (see Figure \ref{fig:uno_pixel_pros}). The ensemble combines 5 cross-validation models with one model trained on the full dataset.

% The base model (SegFormer-B5, pretrained on ADE20K \cite{ade20k}) was trained at 640x640 input resolution using AdamW \cite{adamw} (learning rate of 1e-4 and weight decay of 0.05), batch size 16, and early stopping with patience 10. Key regularization includes label smoothing (0.1) and gradient clipping (max\_norm 1.0), along with augmentations.

% At inference, each model predicts masks which are upsampled to original resolution and combined via pixel-wise averaging. The final mask is obtained by thresholding at 0.5, followed by removal of small regions (area < 100 pixels).

% The main contribution is the ensemble design, which improves performance over a single model by combining cross-validation diversity with a full-data model.

Team UNO Pixel Pros used a SegFormer-B5~\cite{xie2021segformer} segmentation model with a 6-model ensemble (see Figure \ref{fig:uno_pixel_pros}). The ensemble combines 5 cross-validation models with one model trained on the full dataset.

The base model (SegFormer-B5, pretrained on ADE20K~\cite{ade20k}) was trained at 640$\times$640 input resolution using AdamW~\cite{adamw} (learning rate $10^{-4}$ and weight decay 0.05), batch size 16, and early stopping with patience of 10. Regularization included label smoothing (0.1), gradient clipping (max\_norm 1.0), and dropout in the decoder. Data augmentation included ElasticTransform, GridDistortion, ShiftScaleRotate, RGBShift, HueSaturationValue, RandomBrightnessContrast, RandomFog, RandomRain, MotionBlur, and CoarseDropout. 

Since SegFormer-B5 is a semantic segmentation model, each model predicts a binary foreground-background mask rather than instance-separated outputs. During inference, the predicted masks are upsampled to the original resolution and combined via pixel-wise averaging across the six models. The averaged mask is thresholded at 0.5. Instance-level masks are then obtained by applying connected components analysis to the thresholded binary mask, which partitions the foreground prediction into distinct connected regions; each connected region is output as a separate instance mask. Small regions (area $< 100$ pixels) are removed, and the remaining instances are converted to COCO polygon format for submission.

\begin{figure}[t]
    \centering
    \includegraphics[width=\columnwidth]{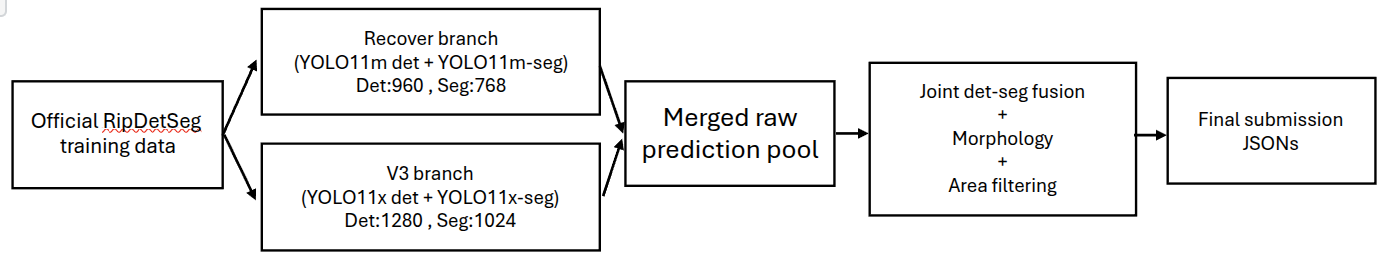}
    \caption{Overview of Team SiGMoid's pipeline. A YOLO11m-based recover branch and a larger YOLO11x-based branch generate detection and segmentation predictions, which are combined in a deterministic fusion pipeline with segmentation refinement, score fusion, weighted box fusion, and area filtering to produce the final predictions.}
    \label{fig:sigmoid_pipeline}
\end{figure}

\begin{figure}[t]
\centering
\includegraphics[width=\linewidth]{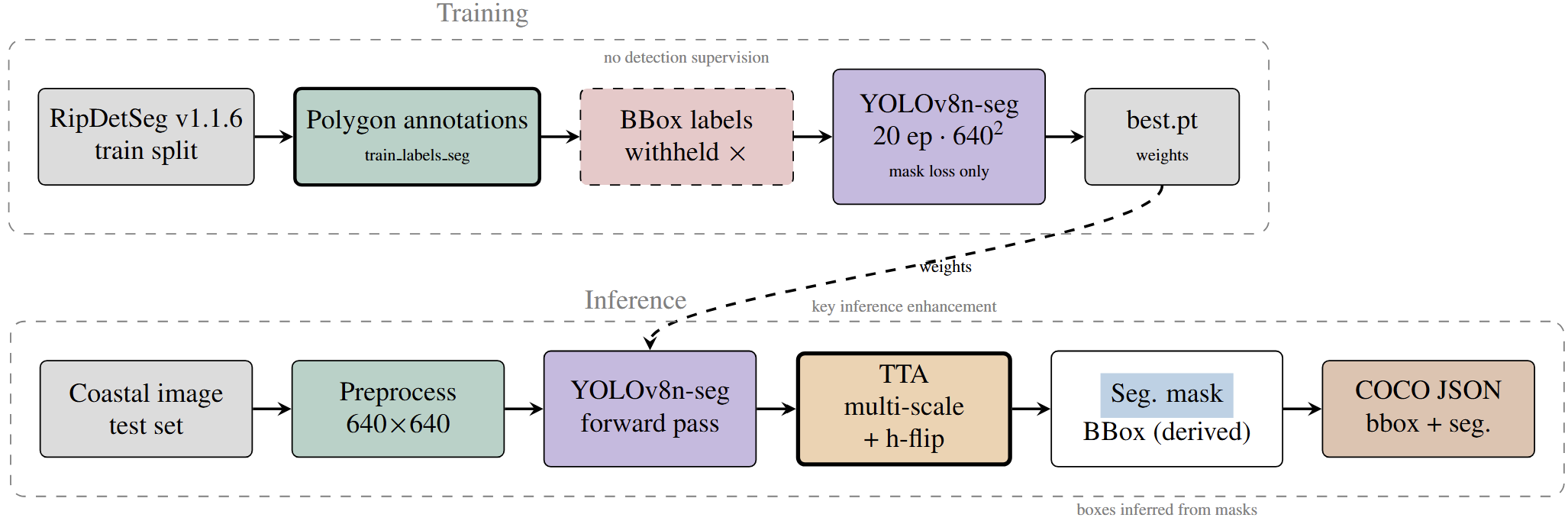}
\caption{Overview of Team Riposte's pipeline. Training only uses segmentation labels; bounding box labels are intentionally withheld.
} % At inference, TTA (multi-scale + horizontal flip) is applied before exporting predictions.
\label{fig:team_riposte}
\end{figure}

\subsection{Team SiGMoid}

Team SiGMoid used a two-branch ensemble with separate YOLO11 \cite{yolo11} detection and segmentation models, combined through a deterministic fusion pipeline (see Figure \ref{fig:sigmoid_pipeline}).

Two model scales were used: a recover branch based on YOLO11m and a second branch based on the larger YOLO11x model. For each branch, detection and segmentation models were trained separately using AdamW with cosine learning-rate scheduling. Training used image sizes from $768$ to $1280$, batch sizes from $3$ to $16$, and a shared augmentation setup including geometric transforms, HSV perturbations, RandAugment \cite{cubuk2020randaugment}, erasing, and stage-dependent mosaic, mixup, and copy-paste. The recover detector was trained in two stages, while the remaining models were trained in a single stage.

At inference, no test-time augmentation was used. Predictions from both branches were generated at fixed scales, with low initial confidence thresholds of $0.03$ for segmentation and $0.01$ for detection. Segmentation polygons were first refined by rasterizing them into binary masks, applying morphological filtering with a single closing iteration using a kernel size of 5, retaining the largest external contour, removing small regions (below an area of $24$ pixels), and simplifying the contour using polygon approximation (with an epsilon ratio of 0.001).

Each refined segmentation prediction was then matched to its single best detection box by IoU, without one-to-one assignment. If the best overlap exceeded $0.20$, the segmentation score was updated as
\[
s' = 0.85\, s_{\text{seg}} + 0.15\, s_{\text{det}},
\]
Otherwise, the segmentation score was reduced by a factor of $0.95$. Segmentation NMS was disabled.

Final detection boxes were produced by weighted box fusion (WBF) \cite{solovyev2021weighted} at IoU $0.45$, applied jointly to raw detection boxes and boxes derived from the refined segmentation predictions from both branches. The confidence thresholds for final filtering were 0.16 for segmentation and 0.18 for detection.
\subsection{Team Riposte}

Team Riposte proposed a lightweight, single-stage approach based on YOLOv8n-seg \cite{yolov8}, using segmentation-only supervision. Instead of training separate detection and segmentation models, they trained only on polygon annotations and intentionally excluded bounding-box labels during training. Instance masks were predicted by the segmentation branch, while detection outputs were obtained at inference by computing axis-aligned bounding boxes from the predicted polygons.

The method follows a segmentation-first design, relying on mask supervision as the only training signal and using the predicted masks to satisfy both the segmentation and detection tasks (see Figure \ref{fig:team_riposte}). Training was performed for 20 epochs at 640$\times$640 resolution with batch size 16, using Adam \cite{adam} and default YOLOv8 hyperparameters. Built-in augmentations, including mosaic, horizontal flipping, and HSV jitter, were retained.

% At inference, test-time augmentation was applied using multiple scales and horizontal flips. Predictions from the augmented views were merged using non-maximum suppression, and the final outputs were exported in COCO-style JSON format with both segmentation polygons and bounding boxes.

\subsection{Team SoloSeg}

\begin{figure}[!t]
  \centering
  \includegraphics[width=\linewidth]{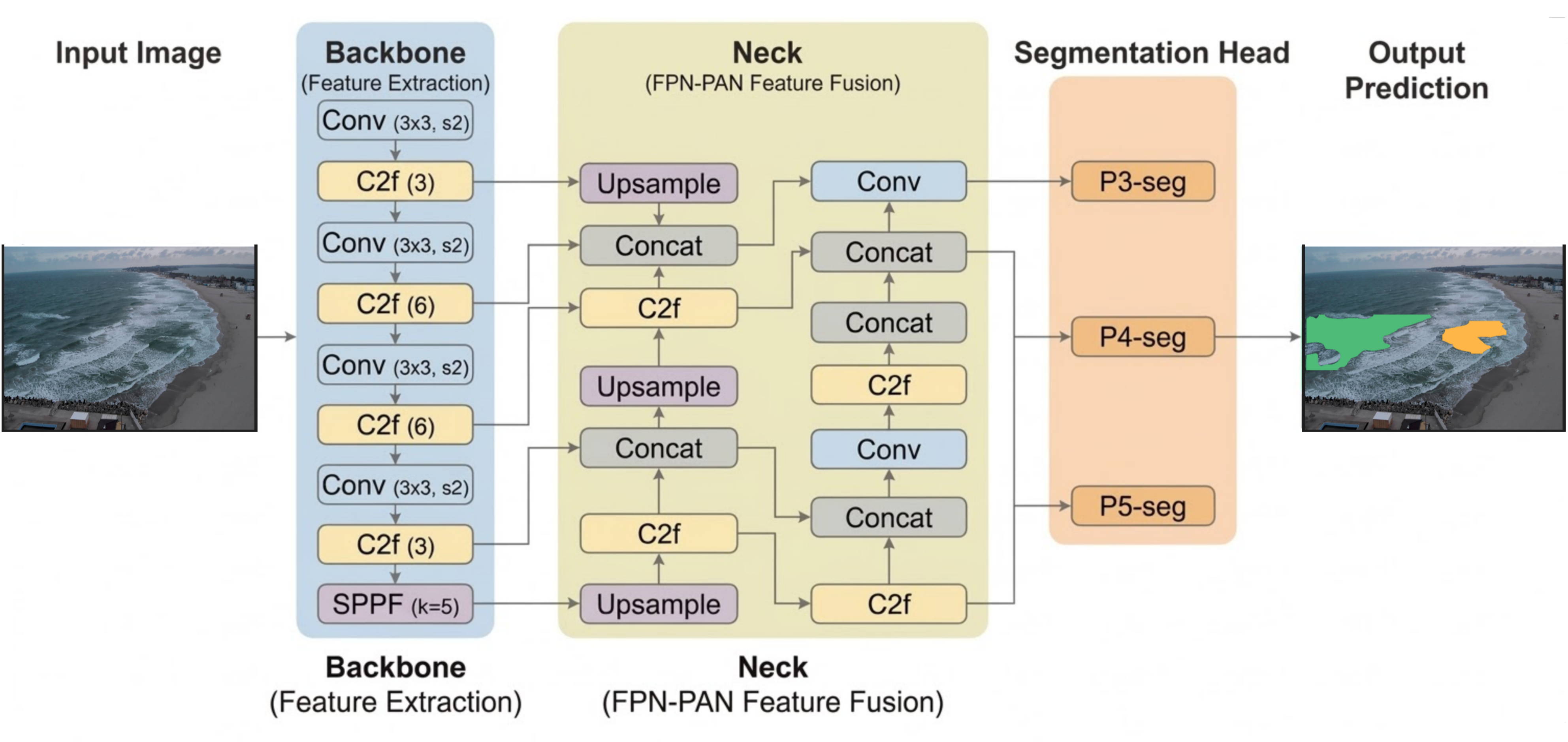}
  \caption{Team Soloseg's schematic illustration of the standard YOLOv8s-seg architecture adopted in the final submission.}
  \label{fig:team_soloseg}
\end{figure}
Team SoloSeg used a YOLOv8s-seg model for rip current instance segmentation, initialized from pretrained COCO-weights and fine-tuned on the training data. Rather than using an ensemble or additional post-processing branches, their final system followed a standard end-to-end segmentation setup with validation-based checkpoint and threshold selection (see Figure \ref{fig:team_soloseg}).

For model development, they created an internal train/validation split of 30{,}934 training images and 1{,}628 validation images, which was used for checkpoint selection and inference tuning. The final model was trained at 1024x1024 resolution for 10 fine-tuning epochs with batch size 32, using AdamW with initial learning rate 0.002, final learning rate factor 0.01, weight decay 0.0005, and standard YOLO augmentations, while turning off the mosaic augmentation for the last 10 epochs.

At inference, they used a confidence threshold of 0.075 and an NMS IoU of 0.40. They further report that the YOLOv8s-seg model trained on the internal split generalized better than the corresponding model trained on the full training set, and this internally validated checkpoint was therefore selected for submission.

% During development, they also evaluated alternative segmentation models and post-processing variants, including YOLOv8n-seg, YOLOv8m-seg, YOLO11s, SparseInst~\cite{cheng2022sparseinst}, morphological opening on predicted masks, stronger augmentation, pseudo-label adaptation on unlabeled validation images, and a CBAM-based ablation~\cite{woo2018cbam}, but none improved over the final single-model YOLOv8s-seg system.

\subsection{Team KMG}

\begin{figure}[t]
    \centering
    \includegraphics[width=\linewidth]{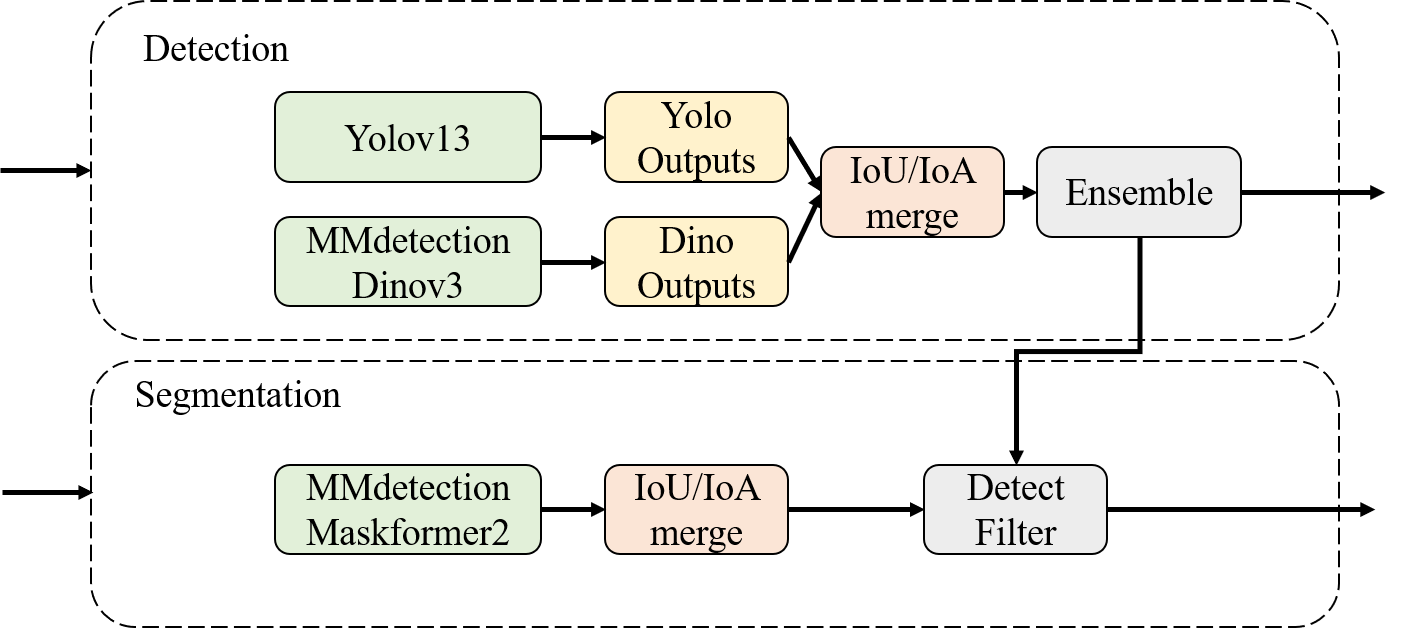}
    \caption{Team KMG's two stage pipeline using YOLOv13 and DINOv3 for detection and Mask2Former with DINOv3 backbone for segmentation.}
    \label{fig:team_kmg}
\end{figure}
Team KMG used a two-stage pipeline in which detection is performed first and then used to refine the segmentation results (see Figure \ref{fig:team_kmg}).

For detection, they ensemble a YOLOv13~\cite{yolo13} detector and a DINOv3-based detector~\cite{dinov3}. The trained YOLOv13 model produces dense predictions of up to $300$ candidates per image. These predictions are post-processed using IoU- and IoA-based merging to combine highly overlapping boxes and suppress redundant detections. In parallel, a DINOv3 detector implemented in MMDetection is applied at high resolution and its output is processed using a similar IoU/IoA-based merging strategy. The two detector outputs are then ensembled by IoU matching: matched boxes are merged by averaging their coordinates and combining their confidence scores, while unmatched high-confidence boxes are retained. 

For segmentation, they use a Mask2Former \cite{mask2former} model with a DINOv3 backbone. At inference, Mask2Former generates instance masks above a confidence threshold, which are then refined in two stages. First, a geometric cleaning and splitting step removes small or noisy fragments and simplifies polygon shapes. Second, a detection-guided refinement step filters and aligns the segmentation masks using the final detection results.

\subsection{Team RIP\_YuvatejaReddy}
\begin{figure}[t]
\centering
\includegraphics[width=\linewidth]{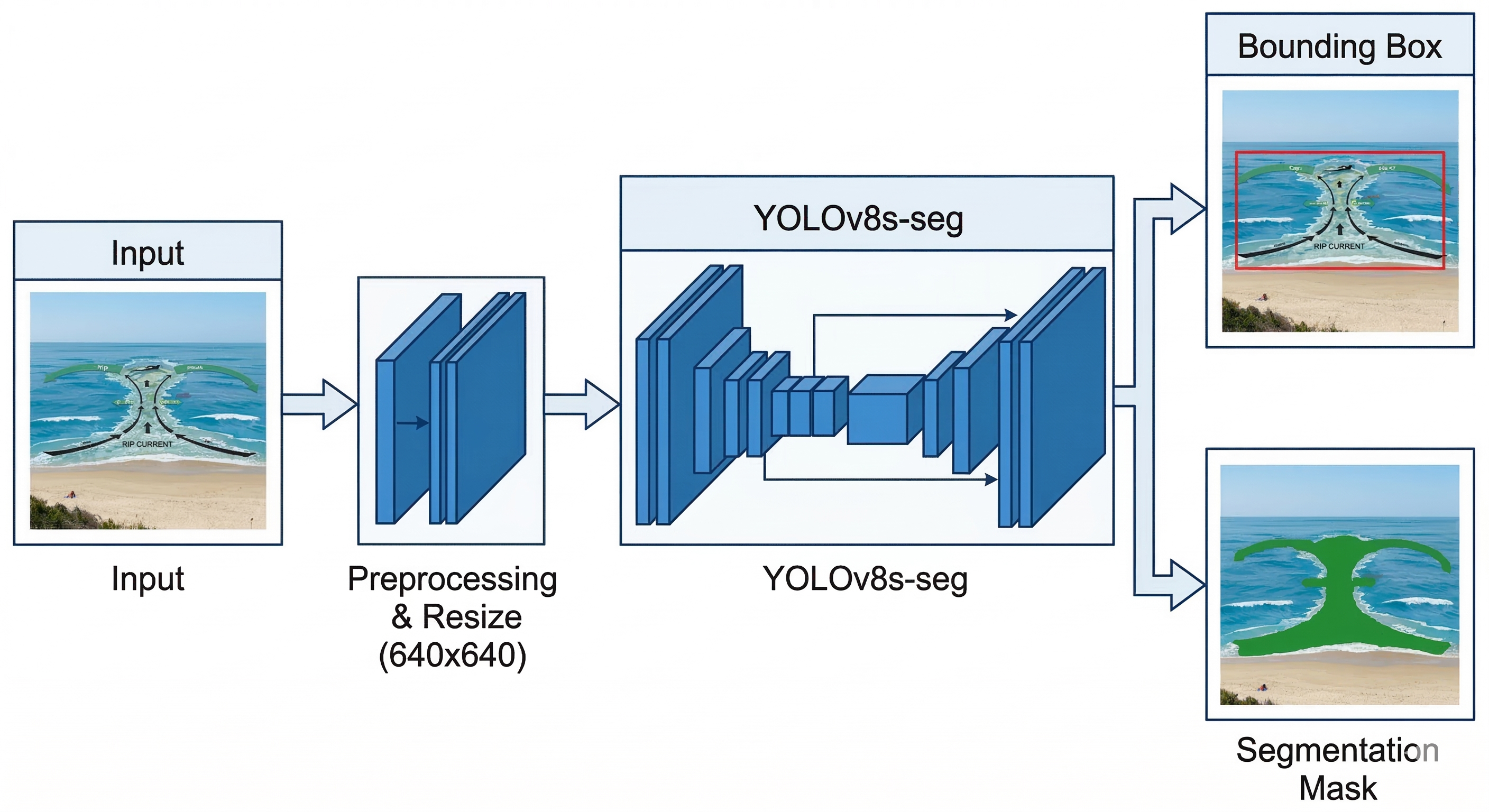}
\caption{Team RIP\_YuvatejaReddy system pipeline.Images are processed by the YOLOv8s-seg, outputting both bounding box coordinates and corresponding polygons.}
\label{fig:team_reddy}
\end{figure}
Team RIP\_YuvatejaReddy used a single YOLOv8s-seg model to jointly perform detection and segmentation (see Figure \ref{fig:team_reddy}).

The model (initialized from COCO-pretrained weights) was trained on the official dataset using a $90/10$ split for validation. Training was performed for $33$ epochs at 640x640 with batch size $4$, default YOLO hyperparameters, and an early stopping with patience of $10$.

\subsection{Team VisionX}

\vspace{-6pt}
\begin{figure}[t]
    \centering
    \includegraphics[width=\columnwidth, trim=40 30 40 10, clip]{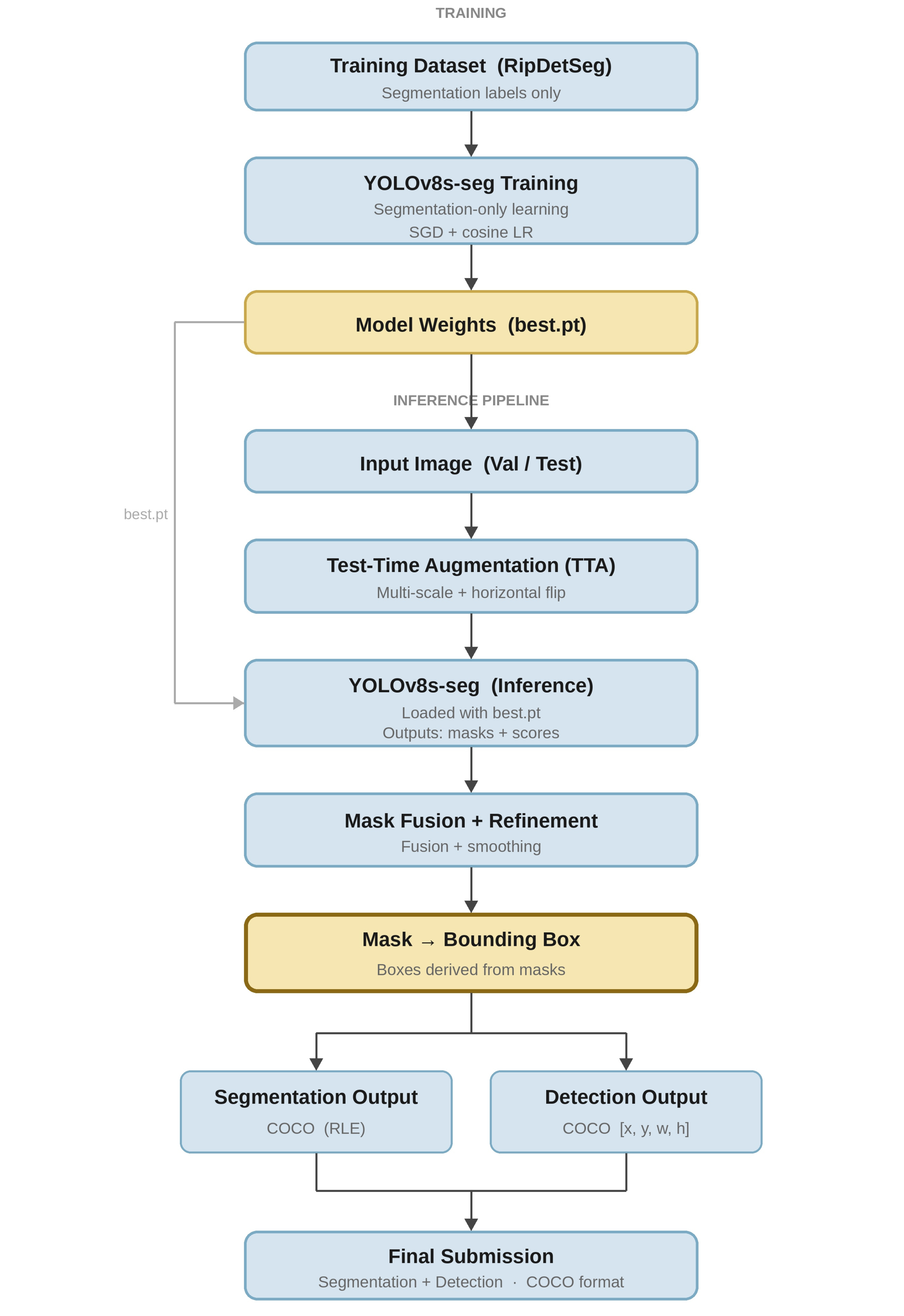}
    \caption{Team VisionX's mask-centric pipeline. Multi-scale inference improves robustness, while fusion and refinement ensure spatial consistency and clean boundaries.}
    \label{fig:team_visionx}
\end{figure}
Team VisionX adopted a mask-centric approach where segmentation is the primary prediction and detection boxes are derived directly from masks (see Figure \ref{fig:team_visionx}).

Their method is based on a YOLOv8s-seg model trained without explicit detection supervision. Bounding boxes are computed from mask extents, enforcing strict spatial alignment and avoiding inconsistencies from separate box regression.

At inference, they use multi-scale evaluation with aggregation across augmented views. Predictions are merged using overlap-based mask fusion (soft merging), followed by morphological refinement to remove noise and improve boundary quality.

The approach tightly couples segmentation and detection, relying entirely on mask quality to drive both tasks.

\subsection{Team NTR}

\begin{figure}[t]
    \centering
    \includegraphics[width=\linewidth]{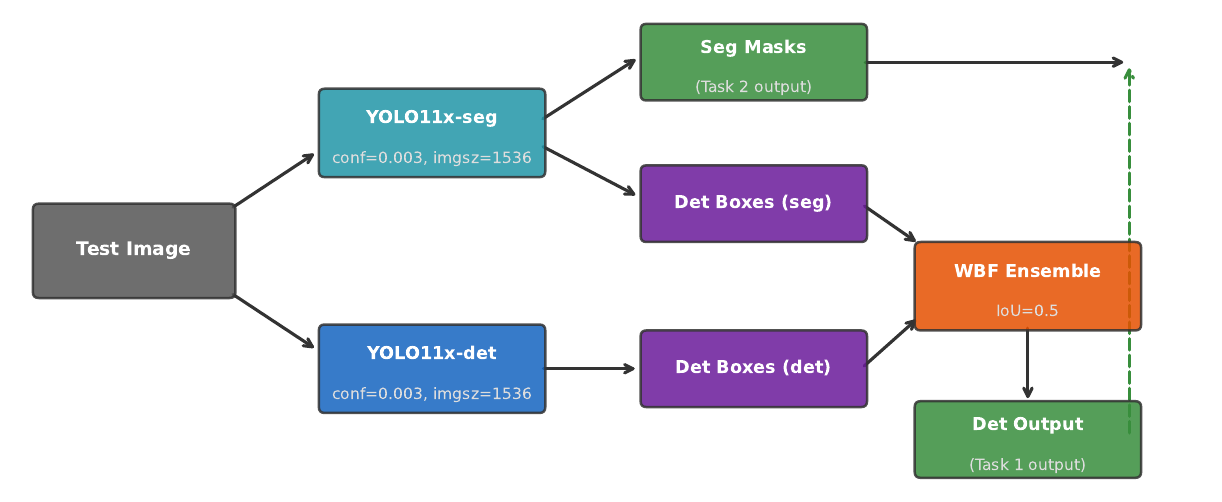}
    \caption{Team NTR's inference pipeline. YOLO11x detection and segmentation models run independently at low confidence. Detected bounding boxes from both are fused via WBF, while segmentation masks come only from the YOLO11x-seg model.}
    \label{fig:ntr_pipeline}
\end{figure}
Team NTR used a two-model YOLO11x ensemble combining a detection model (YOLO11x-det) and a segmentation model (YOLO11x-seg) (see Figure \ref{fig:ntr_pipeline}), with predictions fused via WBF.

The detection model was trained for $150$ epochs at $1280\times1280$ (batch size $16$, SGD with learning rate $0.001$, momentum $0.937$, weight decay $5e-4$), while the segmentation model was trained for $80$ epochs at $640\times640$ with the same optimizer. Both models were initialized from pretrained weights and trained only on the official dataset.

At inference, both models run at $1536\times1536$ with a very low confidence threshold ($0.003$) and NMS IoU $0.6$ to maximize recall. Bounding boxes from both models are fused using WBF (IoU $0.5$, equal weights). Segmentation masks are taken from the segmentation model and post-processed using morphological opening ($5\times5$ kernel, $3$ iterations).

The approach combines high-resolution inference, aggressive recall tuning, and model ensembling to improve detection and segmentation performance.

\subsection{Team Amitabh}
\begin{figure}[t]
    \centering
    % Replace this with your own pipeline figure if available
    \includegraphics[width=1\columnwidth]{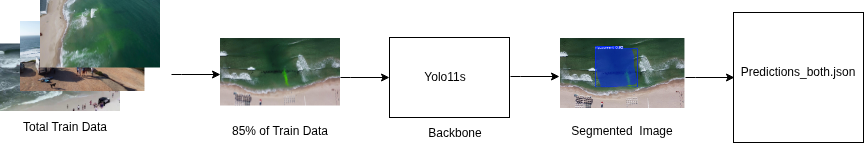}
    \caption{Overview of Team Amitabh’s YOLO11s-seg pipeline.}
    \label{fig:team_amitabh}
\end{figure}

Team Amitabh used a single YOLO11s segmentation model to jointly predict bounding boxes and instance masks (see Figure \ref{fig:team_amitabh}).

The model was trained on the official dataset using an $85/15$ train-validation split, with input resolution $1024$, batch size $8$, and $200$ epochs. Training follows a standard YOLO segmentation setup with box, segmentation, classification, and Distribution Focal Loss(DFL) \cite{dfl} loss.

Inference uses standard YOLO outputs with a confidence threshold of 0.05 and NMS IoU threshold of $0.7$. No ensembling or post-processing is applied.

\section{Conclusion}
This report presented the NTIRE 2026 Rip Current Detection and Segmentation (RipDetSeg) Challenge, providing a detailed overview of the challenge format, the ranking, and insights into the 9 final submissions, evaluated across $F_1[50]$, $F_1[40\!:\!95]$, $F_2[50]$, $F_2[40\!:\!95]$ in the context of axis-aligned bounding box detection and instance segmentation. The submitted solutions relied more on pretrained detection and segmentation models, especially YOLO-based instance segmentation, combined with augmentation and fine-tuning on the challenge data. Performance gains were often driven less by radically new architectures and more by inference-time design, including ensembling, mask refinement, and geometry-aware post-processing to improve spatial consistency.  The final leaderboard, summarized in Tables ~\ref{tab:test_results_segmentation} and \ref{tab:test_results_detection}, shows the difficulty of rip current detection and segmentation: even the top-performing method achieved a computed score of only $55.79$ for segmentation and $56.33$ for detection. Recent solutions, especially YOLO-based ones, are easy to use and efficient. These results suggest that while rip current understanding benefits strongly from progress in robust general-purpose vision models, there is ample room for improvement through future methods tailored to the unique structure of coastal flow patterns. 

Looking ahead, future editions of this challenge may explore a broader range of model architectures, the incorporation of temporal information from video, and the integration of multi-modal data or contextual signals relevant to rip current formation. Increased participation, together with the release of strong open baselines, has the potential to further accelerate progress in this domain. By continuously refining both the dataset and the evaluation protocol, the challenge aims to drive the development of more accurate, robust, and deployable solutions for real-world beach safety applications.

\section*{Acknowledgments}
This research is supported by the project ``Romanian Hub for Artificial Intelligence--HRIA'', Smart Growth, Digitization and Financial Instruments Program, 2021-2027, MySMIS no.~351416. 

This work was partially supported by the Humboldt Foundation. We thank the University of Bucharest and NTIRE 2026 sponsors: OPPO, Kuaishou, and the University of W\"urzburg (Computer Vision Lab).

{
    \small
    \bibliographystyle{ieeenat_fullname}
    \bibliography{main}

@String(CVPR= {IEEE Conf. Comput. Vis. Pattern Recog.})

@String(ICCV= {Int. Conf. Comput. Vis.})

@String(ECCV= {Eur. Conf. Comput. Vis.})

@String(AAAI = {AAAI})

@String(CVPR  = {CVPR})

@String(ICCV  = {ICCV})

@String(ECCV  = {ECCV})

@article{khan2025ripfinder,
  title="{RipFinder: Real-time rip current detection on mobile devices}",
  author={Khan, Fahim and De Silva, Akila and Palinkas, Ashleigh and Dusek, Gregory and Davis, James and Pang, Alex},
  journal={Frontiers in Marine Science},
  volume={12},
  pages={1549513},
  year={2025},
  publisher={Frontiers Media SA}
}

@article{khan2025ripscout,
  title={RipScout: Realtime ML-Assisted Rip Current Detection and Automated Data Collection Using UAVs},
  author={Khan, Fahim and Stewart, Donald and de Silva, Akila and Palinkas, Ashleigh and Dusek, Gregory and Davis, James and Pang, Alex},
  journal={IEEE Journal of Selected Topics in Applied Earth Observations and Remote Sensing},
  year={2025},
  publisher={IEEE}
}

@article{codabench,
    title = {Codabench: Flexible, easy-to-use, and reproducible meta-benchmark platform},
    author = {Zhen Xu and Sergio Escalera and Adrien Pavão and Magali Richard and 
                Wei-Wei Tu and Quanming Yao and Huan Zhao and Isabelle Guyon},
    journal = {Patterns},
    volume = {3},
    number = {7},
    pages = {100543},
    year = {2022},
    issn = {2666-3899},
    doi = {https://doi.org/10.1016/j.patter.2022.100543},
    url = {https://www.sciencedirect.com/science/article/pii/S2666389922001465}
}

@inproceedings{qian2025ripgan,
    author = {Qian, Shenyang and Harley, Mitchell and Razzak, Imran and Song, Yang},
    title = "{RipGAN: A GAN-based rip current data augmentation method}",
    booktitle = {Proceedings of the IEEE International Conference on Robotics and Automation},
    year={2025}
}

@InProceedings{Dumitriu_2025_CVPR,
    author    = {Dumitriu, Andrei and Tatui, Florin and Miron, Florin and Ralhan, Aakash and Ionescu, Radu Tudor and Timofte, Radu},
    title     = {RipVIS: Rip Currents Video Instance Segmentation Benchmark for Beach Monitoring and Safety},
    booktitle = {Proceedings of the IEEE/CVF Conference on Computer Vision and Pattern Recognition (CVPR)},
    month     = {June},
    year      = {2025},
    pages     = {3427--3437}
}

@article{desilva2021automated,
  title={Automated rip current detection with region based convolutional neural networks},
  author={de Silva, Akila and Mori, Issei and Dusek, Gregory and Davis, James and Pang, Alex},
  journal={Coastal Engineering},
  volume={166},
  pages={103859},
  year={2021},
  note={2, 3}
}

@inproceedings{he2017mask,
  title="{Mask R-CNN}",
  author={He, Kaiming and Gkioxari, Georgia and Doll{\'a}r, Piotr and Girshick, Ross},
  booktitle={Proceedings of the IEEE International Conference on Computer Vision (ICCV)},
  pages={2961--2969},
  year={2017}
}

@inproceedings{mask2former,
  title={Masked-attention mask transformer for universal image segmentation},
  author={Cheng, Bowen and Misra, Ishan and Schwing, Alexander G and Kirillov, Alexander and Girdhar, Rohit},
  booktitle={Proceedings of the IEEE/CVF conference on computer vision and pattern recognition},
  pages={1290--1299},
  year={2022}
}

@inproceedings{choi2024explainable,
  title="{Explainable Rip Current Detection and Visualization with XAI EigenCAM}",
  author={Choi, Juno and Rajendran, Muralidharan and Suh, Yong Cheol},
  booktitle={Proceedings of 26th International Conference on Advanced Communications Technology},
  pages={1--6},
  year={2024},
}

@inproceedings{dumitriu2023rip,
  title="{Rip Current Segmentation: A novel benchmark and YOLOv8 baseline results}",
  author={Dumitriu, Andrei and Tatui, Florin and Miron, Florin and Ionescu, Radu Tudor and Timofte, Radu},
  booktitle={Proceedings of the IEEE/CVF Conference on Computer Vision and Pattern Recognition (CVPR) Workshops},
  pages={1261--1271},
  year={2023}
}

@article{de2023ripviz,
  title="{RipViz: Finding Rip Currents by Learning Pathline Behavior}",
  author={de Silva, Akila and Zhao, Mona and Stewart, Donald and Hasan, Fahim and Dusek, Gregory and Davis, James and Pang, Alex},
  journal={IEEE Transactions on Visualization and Computer Graphics},
  year={2024},
  volume={30},
  number={7},
  pages={3930--3944},
}

@article{rashid2023reducing,
  title={Reducing rip current drowning: An improved residual based lightweight deep architecture for rip detection},
  author={Rashid, Ashraf Haroon and Razzak, Imran and Tanveer, M. and Hobbs, Michael},
  journal={ISA Transactions},
  volume={132},
  pages={199--207},
  year={2023},
}

@article{zhu2022yolo,
  title="{YOLO-Rip: A modified lightweight network for Rip currents detection}",
  author={Zhu, Daoheng and Qi, Rui and Hu, Pengpeng and Su, Qianxin and Qin, Xue and Li, Zhiqiang},
  journal={Frontiers in Marine Science},
  volume={9},
  pages={930478},
  year={2022},
}

@article{mcgill2022flow,
  title={Rip current and channel detection using surfcams and optical flow},
  author={McGill, Sean P. and Ellis, Jean T.},
  journal={Shore \& Beach},
  volume={90},
  number={1},
  pages={50},
  year={2022}
}

@article{mori2022flow,
  title={Flow-Based Rip Current Detection and Visualization},
  author={Mori, Issei and de Silva, Akila and Dusek, Gregory and Davis, James and Pang, Alex},
  journal={IEEE Access},
  volume={10},
  pages={6483--6495},
  year={2022},
}

@article{rampal2022interpretable,
  title={Interpretable deep learning applied to rip current detection and localization},
  author={Rampal, Neelesh and Shand, Tom and Wooler, Adam and Rautenbach, Christo},
  journal={Remote Sensing},
  volume={14},
  number={23},
  pages={6048},
  year={2022},
}

@inproceedings{rashid2021ripdet,
  title="{RipDet: A fast and lightweight deep neural network for rip currents detection}",
  author={Rashid, Ashraf Haroon and Razzak, Imran and Tanveer, Muhammad and Robles-Kelly, Antonio},
  booktitle={Proceedings of 2021 International Joint Conference on Neural Networks},
  pages={1--6},
  year={2021},
}

@inproceedings{rashid2020ripnet,
  title={RipNet: A lightweight one-class deep neural network for the identification of rip currents},
  author={Rashid, Ashraf Haroon and Razzak, Imran and Tanveer, Muhammad and Robles-Kelly, Antonio},
  booktitle={Proceedings of 27th International Conference on Neural Information Processing},
  pages={172--179},
  year={2020},
}

@article{maryan2019machine,
  title={Machine learning applications in detecting rip channels from images},
  author={Maryan, Corey and Hoque, Md Tamjidul and Michael, Christopher and Ioup, Elias and Abdelguerfi, Mahdi},
  journal={Applied Soft Computing},
  volume={78},
  pages={84--93},
  year={2019},
}

@inproceedings {philip2016flow,
booktitle = {Proceedings of the Eurographics / IEEE VGTC Conference on Visualization: Short Papers},
pages = {19--23},
title = {{Detecting and Visualizing Rip Current Using Optical Flow}},
author = {Philip, Shweta and Pang, Alex},
year = {2016},
DOI = {10.2312/eurovisshort.20161155}
}

@article{11_ali2025ibn,
  title={IBN-Driven Rip Current Analysis Using UAVs for Next-Generation Coastal Surveillance},
  author={Ali, Shehzad and Saqib, Muhammad and Saudagar, Abdul Khader Jilani and Sajjad, Muhammad and Hijji, Mohammad and Alkhrijah, Yazeed Masaud and Muhammad, Khan and De Albuquerque, Victor Hugo C},
  journal={IEEE Internet of Things Journal},
  year={2025},
  publisher={IEEE}
}

@article{12_choi2025box2rip,
  title={Box2Rip: Instance Segmentation of Amorphous Rip Currents via Box-Supervised Learning},
  author={Choi, Juno and Rajendran, Muralidharan and Suh, Yong Cheol},
  journal={IEEE Access},
  year={2025},
  publisher={IEEE}
}

@article{29_sun2023rip,
  title={Rip Current Detection in Nearshore Areas through UAV Video Analysis with Almost Local-Isometric Embedding Techniques on Sphere},
  author={Sun, Anchen and Yang, Kaiqi},
  journal={arXiv preprint arXiv:2304.11783},
  year={2023}
}

@inproceedings{80_wan2026ripalert,
  title={RipAlert: A Future-Frame-Aware Framework for Rip Current Forecasting and Early Alerting},
  author={Wan, Meng and Su, Qi and Xia, Zhixin and Chen, Kanglin and Wang, Jue and Liu, Tiantian and Cao, Rongqiang and Cui, Hui and Shi, Peng and Wang, Yangang and others},
  booktitle={Proceedings of the AAAI Conference on Artificial Intelligence},
  volume={40},
  number={46},
  pages={39368--39377},
  year={2026}
}

@article{76_catala2026oriented,
  title={Oriented Object Detection for Complex Hydrodynamic Features: A Multi-Platform Rip Current Identification System},
  author={Catal{\`a}-Gonell, Albert and Soriano-Gonz{\'a}lez, Jes{\'u}s and S{\'a}nchez-Garc{\'\i}a, Elena and Criado-Sudau, Francisco Fabi{\'a}n and Oliver-Sans{\'o}, Josep and Kozlov, Valentin and Alibabaei, Khadijeh and Lisani, Jos{\'e} Luis and Fern{\'a}ndez-Mora, {\`A}ngels},
  journal={EGUsphere},
  volume={2026},
  pages={1--29},
  year={2026},
  publisher={Copernicus Publications G{\"o}ttingen, Germany}
}

@article{72_putri2025optimizing,
  title={Optimizing Data Augmentation Parameters in YOLOv11 for Enhanced Rip Current Detection on Small Datasets from Depok-Parangtritis Coastline},
  author={Putri, Madina Hayva and Zaky, Umar and Prabawa, Bayu Argadyanto},
  journal={Jurnal Teknik Informatika (Jutif)},
  volume={6},
  number={5},
  pages={3938--3957},
  year={2025}
}

@article{75_liu2026deep,
  title={A Deep Learning-Based Pipeline for Detecting Rip Currents from Satellite Imagery},
  author={Liu, Yuli and Yang, Yifei and Li, Xiang and Yang, Fan and Xie, Huarong and Wang, Wei and Dong, Changming},
  journal={Remote Sensing},
  volume={18},
  number={2},
  pages={368},
  year={2026},
  publisher={MDPI}
}

@article{brander2000morphodynamics,
  title="{Morphodynamics of a large-scale rip current system at Muriwai Beach, New Zealand}",
  author={Brander, Robert W. and Short, A.D.},
  journal={Marine Geology},
  volume={165},
  number={1-4},
  pages={27--39},
  year={2000},
}

@misc{vis2021,
  author       = "Linjie Yang and Yuchen Fan and Yang Fu and Ning Xu",
  title        = "The 3rd Large-scale Video Object Segmentation Challenge - video instance segmentation track",
  month        = jun,
  year         = "2021",
}

@inproceedings{ Yang2019vis,
    author = {Linjie Yang and Yuchen Fan and Ning Xu},
    title = {Video Instance Segmentation},
    booktitle={Proceedings of the IEEE/CVF International Conference on Computer Vision (ICCV)},
  pages={5188--5197},
  year={2019}
 }

@inproceedings{cordts2016cityscapes,
  title={The cityscapes dataset for semantic urban scene understanding},
  author={Cordts, Marius and Omran, Mohamed and Ramos, Sebastian and Rehfeld, Timo and Enzweiler, Markus and Benenson, Rodrigo and Franke, Uwe and Roth, Stefan and Schiele, Bernt},
  booktitle={Proceedings of the IEEE Conference on Computer Vision and Pattern Recognition (CVPR)},
  pages={3213--3223},
  year={2016}
}

@inproceedings{lin2014microsoft,
  title="{Microsoft COCO: Common objects in context}",
  author={Lin, Tsung-Yi and Maire, Michael and Belongie, Serge and Hays, James and Perona, Pietro and Ramanan, Deva and Doll{\'a}r, Piotr and Zitnick, C Lawrence},
  booktitle={Proceedings of 13th European conference on Computer Vision (ECCV)},
  pages={740--755},
  year={2014},
  organization={Springer}
}

@article{da2003analysis,
  title="{Analysis of hazards associated with sea bathing: results of five years work in oceanic beaches of Santa Catarina state, southern Brazil}",
  author={Da F. Klein, A.H. and Santana, G.G. and Diehl, F.L. and De Menezes, J.T.},
  journal={Journal of Coastal Research},
  pages={107--116},
  year={2003},
}

@article{lushine1991study,
  title={A study of rip current drownings and related weather factors},
  author={Lushine, James B.},
  journal={National Weather Digest},
  volume={16},
  number={3},
  pages={13--19},
  year={1991}
}

@article{brewster2019estimations,
  title="{Estimations of rip current rescues and drowning in the United States}",
  author={Brewster, B. Chris and Gould, Richard E. and Brander, Robert W.},
  journal={Natural Hazards and Earth System Sciences},
  volume={19},
  number={2},
  pages={389--397},
  year={2019},
}

@article{brander2013brief,
  title="{Brief Communication: A new perspective on the Australian rip current hazard}",
  author={Brander, R. and Dominey-Howes, Dale and Champion, C. and Del Vecchio, O. and Brighton, B.},
  journal={Natural Hazards and Earth System Sciences},
  volume={13},
  number={6},
  pages={1687--1690},
  year={2013},
}

@article{castelle2016rip,
  title={Rip current types, circulation and hazard},
  author={Castelle, B. and Scott, Tim and Brander, R.W. and McCarroll, R.J.},
  journal={Earth-Science Reviews},
  volume={163},
  pages={1--21},
  year={2016},
}

@misc{noaa2023ripcurrents,
  author = {{National Oceanic and Atmospheric Administration (NOAA)}},
  title = {What is a rip current?},
  accessed = {2023},
  howpublished = {https://oceanservice.noaa.gov/facts/ripcurrent.html},
  note = {Accessed: March, 2023},
year={2023}
}

@article{ravi2024sam,
  title="{SAM 2: Segment Anything in Images and Videos}",
  author={Ravi, Nikhila and Gabeur, Valentin and Hu, Yuan-Ting and Hu, Ronghang and Ryali, Chaitanya and Ma, Tengyu and Khedr, Haitham and R{\"a}dle, Roman and Rolland, Chloe and Gustafson, Laura and others},
  journal={arXiv preprint arXiv:2408.00714},
  year={2024}
}

@inproceedings{kirillov2023segment,
  title={Segment anything},
  author={Kirillov, Alexander and Mintun, Eric and Ravi, Nikhila and Mao, Hanzi and Rolland, Chloe and Gustafson, Laura and Xiao, Tete and Whitehead, Spencer and Berg, Alexander C and Lo, Wan-Yen and others},
  booktitle={Proceedings of the IEEE/CVF International Conference on Computer Vision (ICCV)},
  pages={4015--4026},
  year={2023}
}

@inproceedings{aim2025ripseg,
  title={{AIM} 2025 rip current segmentation (RipSeg) challenge report},
  author={Andrei Dumitriu and Florin Miron and Florin Tatui and Radu Tudor Ionescu and Radu Timofte and Aakash Ralhan and Florin-Alexandru Vasluianu and others},
  booktitle={Proceedings of the IEEE/CVF International Conference on Computer Vision (ICCV) Workshops},
  year={2025}
}

@inproceedings{ntire26deepfake, 
title={{    Robust Deepfake Detection, NTIRE 2026 Challenge: Report    }}, 
author={    Hopf, Benedikt and  Timofte, Radu and others    },   
booktitle={Proceedings of the IEEE/CVF Conference on Computer Vision and Pattern Recognition (CVPR) Workshops},  
year = {2026} 
}

@inproceedings{ntire26hrdepth, 
title={{    NTIRE 2026 Challenge on High-Resolution Depth of non-Lambertian Surfaces    }}, 
author={    Zama Ramirez, Pierluigi and  Tosi, Fabio and  Di Stefano, Luigi and  Timofte, Radu and  Costanzino, Alex and  Poggi, Matteo and  Salti, Samuele and  Mattoccia, Stefano and others    },   
booktitle={Proceedings of the IEEE/CVF Conference on Computer Vision and Pattern Recognition (CVPR) Workshops},  
year = {2026} 
}

@inproceedings{ntire26raim_fusion, 
title={{    NTIRE 2026 The 3rd Restore Any Image Model (RAIM) Challenge: Multi-Exposure Image Fusion in Dynamic Scenes (Track2)    }}, 
author={    Qu, Lishen and  Liu, Yao and  Liang, Jie and  Zeng, Hui and  Dai, Wen and  Guan, Ya-nan and  Qin, Guanyi and  Zhou, Shihao and  Yang, Jufeng and  Zhang, Lei and  Timofte, Radu and others    },   
booktitle={Proceedings of the IEEE/CVF Conference on Computer Vision and Pattern Recognition (CVPR) Workshops},  
year = {2026} 
}

@inproceedings{ntire26raim_portrait, 
title={{    NTIRE 2026 The 3rd Restore Any Image Model (RAIM) Challenge: AI Flash Portrait (Track 3)    }}, 
author={    Guan, Ya-nan and  Zhang, Shaonan and  Guo, Hang and  Wang, Yawen and  Fan, Xinying and  Liang, Jie and  Zeng, Hui and  Qin, Guanyi and  Qu, Lishen and  Dai, Tao and  Xia, Shu-Tao and  Zhang, Lei and  Timofte, Radu and others    },   
booktitle={Proceedings of the IEEE/CVF Conference on Computer Vision and Pattern Recognition (CVPR) Workshops},  
year = {2026} 
}

@inproceedings{ntire26raim_piqa, 
title={{    NTIRE 2026 The 3rd Restore Any Image Model (RAIM) Challenge: Professional Image Quality Assessment (Track 1)    }}, 
author={    Qin, Guanyi and  Liang, Jie and  Zhang, Bingbing and  Qu, Lishen and  Guan, Ya-nan and  Zeng, Hui and  Zhang, Lei and  Timofte, Radu and others    },   
booktitle={Proceedings of the IEEE/CVF Conference on Computer Vision and Pattern Recognition (CVPR) Workshops},  
year = {2026} 
}

@inproceedings{ntire26lightsr, 
title={{    NTIRE 2026 Challenge on Light Field Image Super-Resolution: Methods and Results    }}, 
author={    Wang, Yingqian and  Liang, Zhengyu and  Zhang, Fengyuan and  Zhao, Wending and  Wang, Longguang and  Li, Juncheng and  Yang, Jungang and  Timofte, Radu and  Guo, Yulan and others    },   
booktitle={Proceedings of the IEEE/CVF Conference on Computer Vision and Pattern Recognition (CVPR) Workshops},  
year = {2026} 
}

@inproceedings{ntire263dsr, 
title={{    NTIRE 2026 Challenge on 3D Content Super-Resolution: Methods and Results    }}, 
author={    Wang, Longguang and  Guo, Yulan and  Wang, Yingqian and  Li, Juncheng and  Peng, Sida and  Zhang, Ye and  Timofte, Radu and  Chen, Minglin and  Wang, Yi and  Hu, Qibin and  Lei, Wenjie and others    },   
booktitle={Proceedings of the IEEE/CVF Conference on Computer Vision and Pattern Recognition (CVPR) Workshops},  
year = {2026} 
}

@inproceedings{ntire26videores, 
title={{    NTIRE 2026 Challenge on Bitstream-Corrupted Video Restoration: Methods and Results    }}, 
author={    Zou, Wenbin and  Liu, Tianyi and  Wu, Kejun and  Zhuang, Huiping and  Wu, Zongwei and  Zhou, Zhuyun and  Timofte, Radu and  others     },   booktitle={Proceedings of the IEEE/CVF Conference on Computer Vision and Pattern Recognition (CVPR) Workshops},  
year = {2026} 
}

@inproceedings{ntire26XAIGCqa, 
title={{    NTIRE 2026 X-AIGC Quality Assessment Challenge: Methods and Results    }}, 
author={    Liu, Xiaohong and  Min, Xiongkuo and  Zhai, Guangtao and  Hu, Qiang and  Cao, Jiezhang and  Zhou, Yu and  Sun, Wei and  Wen, Farong and  Xu, Zitong and  Zhou, Yingjie and  Duan, Huiyu and  Liu, Lu and  Wang, Jiarui and  Luo, Siqi and  Li, Chunyi and  Xu, Li and  Zhang, Zicheng and  Shi, Yue and  Wang, Yubo and  Zhang, Minghong and  Guo, Chunchao and  Hu, Zhichao and  Chen, Mingtao and  Wu, Xiele and  Ma, Xin and  Lv, Zhaohe and  Xue, Yuanhao and  Wang, Jiaqi and  Sha, Xinxing and  Timofte, Radu and  others    },   
booktitle={Proceedings of the IEEE/CVF Conference on Computer Vision and Pattern Recognition (CVPR) Workshops},  
year = {2026} 
}

@inproceedings{ntire26shadow, 
title={{    Advances in Single-Image Shadow Removal: Results from the NTIRE 2026 Challenge    }}, 
author={    Vasluianu, Florin-Alexandru and  Seizinger, Tim and  Zhou, Zhuyun and  Wu, Zongwei and  Timofte, Radu and  others     },   
booktitle={Proceedings of the IEEE/CVF Conference on Computer Vision and Pattern Recognition (CVPR) Workshops},  
year = {2026} 
}

@inproceedings{ntire26lightnorm, 
title={{    Learning-Based Ambient Lighting Normalization: NTIRE 2026 Challenge Results and Findings    }}, 
author={    Vasluianu, Florin-Alexandru and  Seizinger, Tim and  Chen, Jeffrey and  Zhou, Zhuyun and  Wu, Zongwei and  Timofte, Radu and  others    },   booktitle={Proceedings of the IEEE/CVF Conference on Computer Vision and Pattern Recognition (CVPR) Workshops},  
year = {2026} 
}

@inproceedings{ntire26bokeh, 
title={{    The First Controllable Bokeh Rendering Challenge at NTIRE 2026    }}, 
author={    Seizinger, Tim and  Vasluianu, Florin-Alexandru and  Conde, Marcos V. and  Chen, Jeffrey and  Zhou, Zhuyun and  Wu, Zongwei and  Timofte, Radu and  others    },   
booktitle={Proceedings of the IEEE/CVF Conference on Computer Vision and Pattern Recognition (CVPR) Workshops},  
year = {2026} 
}

@inproceedings{ntire26ripdetseg, 
title={{    NTIRE 2026 Rip Current Detection and Segmentation (RipDetSeg) Challenge Report    }}, 
author={    Dumitriu, Andrei and  Ralhan, Aakash and  Miron, Florin and  Tatui, Florin and  Ionescu, Radu Tudor and  Timofte, Radu and  others     },   booktitle={Proceedings of the IEEE/CVF Conference on Computer Vision and Pattern Recognition (CVPR) Workshops},  
year = {2026} 
}

@inproceedings{ntire26llie, 
title={{    Low Light Image Enhancement Challenge at NTIRE 2026    }}, 
author={    Ciubotariu, George and  S M A,  Sharif and  Rehman, Abdur and  Ali Dharejo, Fayaz and  Naqvi, Rizwan Ali and  Conde, Marcos and  Timofte, Radu and others    },   
booktitle={Proceedings of the IEEE/CVF Conference on Computer Vision and Pattern Recognition (CVPR) Workshops},  
year = {2026} 
}

@inproceedings{ntire26highfps, 
title={{    High FPS Video Frame Interpolation Challenge at NTIRE 2026    }}, 
author={    Ciubotariu, George and  Zhou, Zhuyun and  Jin, Yeying and  Wu, Zongwei and  Timofte, Radu and  others    },   
booktitle={Proceedings of the IEEE/CVF Conference on Computer Vision and Pattern Recognition (CVPR) Workshops},  
year = {2026} 
}

@inproceedings{ntire26nthaze, 
title={{    NT-HAZE: A Benchmark Dataset for Realistic Night-time Image Dehazing    }}, 
author={    Ancuti, Radu and  Ancuti, Codruta and  Timofte, Radu and  Ancuti, Cosmin    },   
booktitle={Proceedings of the IEEE/CVF Conference on Computer Vision and Pattern Recognition (CVPR) Workshops},  
year = {2026} 
}

@inproceedings{ntire26nthaze_rep, 
title={{    NTIRE 2026 Nighttime Image Dehazing Challenge Report    }}, 
author={    Ancuti, Radu and  Brateanu, Alexandru and  Vasluianu, Florin and  Balmez, Raul and  Orhei, Ciprian and  Ancuti, Codruta and  Timofte, Radu and  Ancuti, Cosmin and others    },   
booktitle={Proceedings of the IEEE/CVF Conference on Computer Vision and Pattern Recognition (CVPR) Workshops},  
year = {2026} 
}

@inproceedings{ntire26isp, 
title={{    NTIRE 2026 Challenge on Learned Smartphone ISP with Unpaired Data: Methods and Results    }}, 
author={    Perevozchikov, Georgy and  Vladimirov, Daniil and  Timofte, Radu and  others    },   
booktitle={Proceedings of the IEEE/CVF Conference on Computer Vision and Pattern Recognition (CVPR) Workshops},  
year = {2026} 
}

@inproceedings{ntire26ugcvideo, 
title={{    NTIRE 2026 Challenge on Short-form UGC Video Restoration in the Wild with Generative Models: Datasets, Methods and Results    }}, author={    Li, Xin and  Gong, Jiachao and  Wang, Xijun and  Xiong, Shiyao and  Li, Bingchen and  Yao, Suhang  and  Zhou, Chao and  Chen, Zhibo and  Timofte, Radu and others    },   
booktitle={Proceedings of the IEEE/CVF Conference on Computer Vision and Pattern Recognition (CVPR) Workshops},  
year = {2026} 
}

@inproceedings{ntire26dual_focus, 
title={{    NTIRE 2026 The Second Challenge on Day and Night Raindrop Removal for Dual-Focused Images: Methods and Results    }}, 
author={    Li, Xin and  Jin, Yeying and  Yao, Suhang and  Lin, Beibei and  Fan, Zhaoxin and   Yan, Wending and  Jin, Xin and  Wu, Zongwei  and  Li, Bingchen  and  Shi, Peishu and  Yang, Yufei and  Li, Yu and  Chen, Zhibo  and  Wen, Bihan and  Tan, Robby and  Timofte, Radu and others    },   
booktitle={Proceedings of the IEEE/CVF Conference on Computer Vision and Pattern Recognition (CVPR) Workshops},  
year = {2026} 
}

@inproceedings{ntire26srx4, 
title={{    The Fourth Challenge on Image Super-Resolution (×4) at NTIRE 2026: Benchmark Results and Method Overview    }}, 
author={    Chen, Zheng and  Liu, Kai and  Wang, Jingkai and  Yan, Xianglong and  Li, Jianze and  Zhang, Ziqing and  Gong, Jue and  Li, Jiatong and  Sun, Lei and  Liu, Xiaoyang and  Timofte, Radu and  Zhang, Yulun and others    },   
booktitle={Proceedings of the IEEE/CVF Conference on Computer Vision and Pattern Recognition (CVPR) Workshops},  
year = {2026} 
}

@inproceedings{ntire26retouching, 
title={{    Photography Retouching Transfer, NTIRE 2026 Challenge: Report    }}, 
author={    Elezabi, Omar and  V. Conde, Marcos and  Wu, Zongwei and  Jin, Yeying and  Timofte, Radu and others    },   
booktitle={Proceedings of the IEEE/CVF Conference on Computer Vision and Pattern Recognition (CVPR) Workshops},  
year = {2026} 
}

@inproceedings{ntire26rwsr, 
title={{    The First Challenge on Mobile Real-World Image Super-Resolution at NTIRE 2026: Benchmark Results and Method Overview    }}, 
author={    Li, Jiatong and  Chen, Zheng and  Liu, Kai and  Wang, Jingkai and  Zhou, Zihan and  Liu, Xiaoyang and  Zhu, Libo and  Timofte, Radu and  Zhang, Yulun and others    },   
booktitle={Proceedings of the IEEE/CVF Conference on Computer Vision and Pattern Recognition (CVPR) Workshops},  
year = {2026} 
}

@inproceedings{ntire26rsirsr, 
title={{    The First Challenge on Remote Sensing Infrared Image Super-Resolution at NTIRE 2026: Benchmark Results and Method Overview    }}, author={    Liu, Kai and  Yue, Haoyang and  Lin, Zeli and  Chen, Zheng and  Wang, Jingkai and  Gong, Jue and  Timofte, Radu and  Zhang, Yulun and  others    },   
booktitle={Proceedings of the IEEE/CVF Conference on Computer Vision and Pattern Recognition (CVPR) Workshops},  
year = {2026} 
}

@inproceedings{ntire26aigendet, 
title={{    NTIRE 2026 Challenge on Robust AI-Generated Image Detection in the Wild    }}, 
author={    Gushchin, Aleksandr and  Abud, Khaled and  Shumitskaya, Ekaterina and  Filippov, Artem and  Bychkov, Georgii and  Lavrushkin, Sergey and  Erofeev, Mikhail and  Antsiferova, Anastasia and  Chen, Changsheng and  Tan, Shunquan and  Timofte, Radu and  Vatolin, Dmitriy and others    },
booktitle={Proceedings of the IEEE/CVF Conference on Computer Vision and Pattern Recognition (CVPR) Workshops},  
year = {2026} 
}

@inproceedings{ntire26cdfsod, 
title={{    The Second Challenge on Cross-Domain Few-Shot Object Detection at NTIRE 2026: Methods and Results    }}, 
author={    Qiu, Xingyu and  Fu, Yuqian and  Geng, Jiawei and  Ren, Bin and  Pan, Jiancheng and  Wu, Zongwei and  Tang, Hao and  Fu, Yanwei and  Timofte, Radu and  Sebe, Nicu and  Elhoseiny, Mohamed and others    },   
booktitle={Proceedings of the IEEE/CVF Conference on Computer Vision and Pattern Recognition (CVPR) Workshops},  
year = {2026} 
}

@inproceedings{ntire26finrec, 
title={{    NTIRE 2026 Challenge on End-to-End Financial Receipt Restoration and Reasoning from Degraded Images: Datasets, Methods and Results    }}, author={    Guan, Bochen and  Li, Jinlong and  Yang, Kangning and  Ke, Chuang and  Cai, Jie and  Vasluianu, Florin and  Timofte, Radu and others    },   booktitle={Proceedings of the IEEE/CVF Conference on Computer Vision and Pattern Recognition (CVPR) Workshops},  
year = {2026} 
}

@inproceedings{ntire26faceres, 
title={{    The Second Challenge on Real-World Face Restoration at NTIRE 2026: Methods and Results    }}, 
author={    Wang, Jingkai and  Gong, Jue and  Chen, Zheng and  Liu, Kai and  Li, Jiatong and  Zhang, Yulun and  Timofte, Radu and  others    },
booktitle={Proceedings of the IEEE/CVF Conference on Computer Vision and Pattern Recognition (CVPR) Workshops},  
year = {2026} 
}

@inproceedings{ntire26reflection, 
title={{    NTIRE 2026 Challenge on Single Image Reflection Removal in the Wild: Datasets, Results, and Methods    }}, 
author={    Cai, Jie and  Yang, Kangning and  Li, Zhiyuan and  Vasluianu, Florin and  Timofte, Radu and others    },   
booktitle={Proceedings of the IEEE/CVF Conference on Computer Vision and Pattern Recognition (CVPR) Workshops},  
year = {2026} 
}

@inproceedings{ntire26anomalydet, 
title={{    NTIRE 2026  Challenge Report on Anomaly Detection of Face Enhancement for UGC Images    }}, 
author={    Zhong, Yan and   Ma,  Qiufang and  Wang, Zhen and  Jiang, Tingting and  Timofte, Radu and others    },   
booktitle={Proceedings of the IEEE/CVF Conference on Computer Vision and Pattern Recognition (CVPR) Workshops},  
year = {2026} 
}

@inproceedings{ntire26videosal, 
title={{    NTIRE 2026 Challenge on Video Saliency Prediction: Methods and Results    }}, 
author={    Moskalenko, Andrey and  Bryncev, Alexey and  Kosmynin, Ivan and  Shilovskaya, Kira and  Erofeev, Mikhail and  Vatolin, Dmitry and  Timofte, Radu and others    },   
booktitle={Proceedings of the IEEE/CVF Conference on Computer Vision and Pattern Recognition (CVPR) Workshops},  
year = {2026} 
}

@inproceedings{ntire26effsr, 
title={{    The Eleventh NTIRE 2026 Efficient Super-Resolution Challenge Report    }}, 
author={    Ren, Bin and  Guo, Hang and  Shu, Yan and  Ma, Jiaqi and  Cui, Ziteng and  Liu, Shuhong  and  Mei, Guofeng  and  Sun, Lei and  Wu, Zongwei and  Khan, Fahad Shahbaz and  Khan, Salman and  Timofte, Radu and  Li, Yawei and others    },   
booktitle={Proceedings of the IEEE/CVF Conference on Computer Vision and Pattern Recognition (CVPR) Workshops},  
year = {2026} 
}

@inproceedings{ntire26realx3d, 
title={{    3D Restoration and Reconstruction in Adverse Conditions: RealX3D Challenge Results    }}, 
author={    Liu, Shuhong and  Cui, Ziteng and  Bao, Chenyu and  Chu, Xuangeng and  Gu, Lin and  Ren, Bin and  Timofte, Radu and  Conde, Marcos V. and others    },   
booktitle={Proceedings of the IEEE/CVF Conference on Computer Vision and Pattern Recognition (CVPR) Workshops},  
year = {2026} 
}

@inproceedings{ntire26denoising, 
title={{    The Third Challenge on Image Denoising at NTIRE 2026: Methods and Results    }}, 
author={    Sun, Lei and  Guo, Hang and  Ren, Bin and  Su, Shaolin and  Wang, Xian and  Pani Paudel, Danda and  Van Gool, Luc and  Timofte, Radu and  Li, Yawei and others    },   
booktitle={Proceedings of the IEEE/CVF Conference on Computer Vision and Pattern Recognition (CVPR) Workshops},  
year = {2026} 
}

@inproceedings{ntire26aberration, 
title={{    NTIRE 2026 The First Challenge on Blind Computational Aberration Correction: Methods and Results    }}, 
author={    Sun, Lei and  Qian, Xiaolong and  Jiang, Qi and  Wang, Xian and  Gao, Yao and  Yang, Kailun and  Wang, Kaiwei and  Timofte, Radu and  Pani Paudel, Danda and  Van Gool, Luc and others    },   
booktitle={Proceedings of the IEEE/CVF Conference on Computer Vision and Pattern Recognition (CVPR) Workshops},  
year = {2026} 
}

@inproceedings{ntire26eventblurr, 
title={{    The Second Challenge on Event-Based Image Deblurring at NTIRE 2026: Methods and Results    }}, 
author={    Sun, Lei and  Li, Weilun and  Wang, Xian and  Li, Zhendong and  Shi, Letian and  Xu, Dannong and  Zhang, Deheng and  Hu, Mengshun and  Guo, Shuang and  Su, Shaolin and  Timofte, Radu and  Pani Paudel, Danda and  Van Gool, Luc and others    },   
booktitle={Proceedings of the IEEE/CVF Conference on Computer Vision and Pattern Recognition (CVPR) Workshops},  
year = {2026} 
}

@inproceedings{ntire26bursthdr, 
title={{    NTIRE 2026 Challenge on Efficient Burst HDR and Restoration: Datasets, Methods, and Results    }}, 
author={    Park, Hyunhee and  Park, Eunpil and  Lee, Sangmin and  Timofte, Radu and others    },   
booktitle={Proceedings of the IEEE/CVF Conference on Computer Vision and Pattern Recognition (CVPR) Workshops},  
year = {2026} 
}

@inproceedings{ntire26twilight, 
title={{    NTIRE 2026 Low-light Enhancement: Twilight Cowboy Challenge    }}, 
author={    Khalin, Aleksei and  Ershov, Egor and  Panshin, Artem and  Korchagin, Sergey and  Lobarev, Georgiy and  Terekhin, Arseniy and  Dorogova, Sofiia and  Shamsutdinov, Amir and  Mamedov, Yasin and  Khalfin, Bakhtiyar and  Sheludko, Bogdan and  Zilyaev, Emil and  Banić, Nikola and  Perevozchikov, Georgy and  Timofte, Radu and others    },   
booktitle={Proceedings of the IEEE/CVF Conference on Computer Vision and Pattern Recognition (CVPR) Workshops},  
year = {2026} 
}

@inproceedings{ntire26effllie, 
title={{    Efficient Low Light Image Enhancement: NTIRE 2026 Challenge Report    }}, 
author={    Yan, Jiebin  and  Tu, Chenyu  and  Lin, Qinghua and  WU, Zongwei and  Zhang , Weixia and  Wang, Zhihua and  Cao, Peibei and  Fang, Yuming  and  Liu, Xiaoning  and  Zhou, Zhuyun and  Timofte, Radu  and  others    },   
booktitle={Proceedings of the IEEE/CVF Conference on Computer Vision and Pattern Recognition (CVPR) Workshops},  
year = {2026} 
}

@inproceedings{xie2021segformer,
 author = {Xie, Enze and Wang, Wenhai and Yu, Zhiding and Anandkumar, Anima and Alvarez, Jose M. and Luo, Ping},
 booktitle = {Advances in Neural Information Processing Systems},
 editor = {M. Ranzato and A. Beygelzimer and Y. Dauphin and P.S. Liang and J. Wortman Vaughan},
 pages = {12077--12090},
 publisher = {Curran Associates, Inc.},
 title = {SegFormer: Simple and Efficient Design for Semantic Segmentation with Transformers},
 url = {https://proceedings.neurips.cc/paper_files/paper/2021/file/64f1f27bf1b4ec22924fd0acb550c235-Paper.pdf},
 volume = {34},
 year = {2021}
}

@inproceedings{cubuk2020randaugment,
  title={Randaugment: Practical automated data augmentation with a reduced search space},
  author={Cubuk, Ekin D and Zoph, Barret and Shlens, Jonathon and Le, Quoc V},
  booktitle={Proceedings of the IEEE/CVF conference on computer vision and pattern recognition workshops},
  pages={702--703},
  year={2020}
}

@article{adam,
  title={Adam: A method for stochastic optimization},
  author={Kingma, Diederik P and Ba, Jimmy},
  journal={arXiv preprint arXiv:1412.6980},
  year={2014}
}

@article{solovyev2021weighted,
  title={Weighted boxes fusion: Ensembling boxes from different object detection models},
  author={Solovyev, Roman and Wang, Weimin and Gabruseva, Tatiana},
  journal={Image and Vision Computing},
  volume={107},
  pages={104117},
  year={2021},
  publisher={Elsevier}
}

@software{yolov8,
  author = {Glenn Jocher and Ayush Chaurasia and Jing Qiu},
  title = {Ultralytics YOLOv8},
  version = {8.0.0},
  year = {2023},
  url = {https://github.com/ultralytics/ultralytics},
  orcid = {0000-0001-5950-6979, 0000-0002-7603-6750, 0000-0003-3783-7069},
  license = {AGPL-3.0}
}

@software{yolo11,
  author = {Glenn Jocher and Jing Qiu},
  title = {Ultralytics YOLO11},
  version = {11.0.0},
  year = {2024},
  url = {https://github.com/ultralytics/ultralytics},
  orcid = {0000-0001-5950-6979, 0000-0003-3783-7069},
  license = {AGPL-3.0}
}

@inproceedings{ade20k,
  title={Scene parsing through ade20k dataset},
  author={Zhou, Bolei and Zhao, Hang and Puig, Xavier and Fidler, Sanja and Barriuso, Adela and Torralba, Antonio},
  booktitle={Proceedings of the IEEE conference on computer vision and pattern recognition},
  pages={633--641},
  year={2017}
}

@article{yolo13,
  title={Yolov13: Real-time object detection with hypergraph-enhanced adaptive visual perception},
  author={Lei, Mengqi and Li, Siqi and Wu, Yihong and Hu, Han and Zhou, You and Zheng, Xinhu and Ding, Guiguang and Du, Shaoyi and Wu, Zongze and Gao, Yue},
  journal={arXiv preprint arXiv:2506.17733},
  year={2025}
}

@article{adamw,
  title={Decoupled weight decay regularization},
  author={Loshchilov, Ilya and Hutter, Frank},
  journal={arXiv preprint arXiv:1711.05101},
  year={2017}
}

@article{dinov3,
  title={DINOv3},
  author={Sim{\'e}oni, Oriane and Vo, Huy V and Seitzer, Maximilian and Baldassarre, Federico and Oquab, Maxime and Jose, Cijo and Khalidov, Vasil and Szafraniec, Marc and Yi, Seungeun and Ramamonjisoa, Micha{\"e}l and others},
  journal={arXiv preprint arXiv:2508.10104},
  year={2025}
}

@article{dfl,
  title={Generalized focal loss: Learning qualified and distributed bounding boxes for dense object detection},
  author={Li, Xiang and Wang, Wenhai and Wu, Lijun and Chen, Shuo and Hu, Xiaolin and Li, Jun and Tang, Jinhui and Yang, Jian},
  journal={Advances in neural information processing systems},
  volume={33},
  pages={21002--21012},
  year={2020}
}
}

\end{document}